\documentclass[acmlarge]{acmart}
\usepackage[superscript]{cite}
\usepackage{algpseudocode}%
\usepackage{array}
\usepackage{tabularx}
\usepackage{subcaption}
\usepackage{caption}
\usepackage[ruled, vlined, linesnumbered]{algorithm2e}
\usepackage{xcolor}
\usepackage{cite}

\usepackage{amsmath,amssymb,amsfonts}
\usepackage{graphicx}
\usepackage{textcomp}
\usepackage[para]{footmisc}
\usepackage{url}
\usepackage{longtable}
\usepackage{pifont}

\usepackage{multirow}
\usepackage[table]{xcolor}
\usepackage{array, graphicx, booktabs, longtable}
\usepackage[table]{xcolor}
\usepackage{breakcites}
\usepackage{ragged2e}
\usepackage{tikz}
\usetikzlibrary{trees, shapes, positioning}
\usetikzlibrary{trees, arrows.meta}
\usetikzlibrary{mindmap,trees}
\usepackage{adjustbox}
\usepackage[edges]{forest}
\usepackage{xcolor}
\usepackage{longtable, array, booktabs, xcolor, ragged2e}

\usepackage[utf8]{inputenc}
\usepackage{tikz}
\usetikzlibrary{arrows.meta, positioning, shapes.multipart, fit, backgrounds}
\usepackage{adjustbox}
\usepackage{xcolor}
\usepackage{lmodern}
\usepackage{caption}
\usepackage[T1]{fontenc}
\usepackage{lmodern}  

\AtBeginDocument{%
  }

\begin{document}

\title{A Survey on Improving Human Robot Collaboration through Vision-and-Language Navigation}




\author{Nivedan Yakolli}
\affiliation{
  \department{Department of Computer Science and Information Systems}
  \institution{Birla Institute of Technology and Science (BITS)}
  \city{Pilani}
  \country{India}
}
\email{p20230032@pilani.bits-pilani.ac.in}

\author{Avinash Gautam}
\affiliation{
  \department{Department of Computer Science and Information Systems}
  \institution{Birla Institute of Technology and Science (BITS)}
  \city{Pilani}
  \country{India}
}
\email{avinash@pilani.bits-pilani.ac.in}

\author{Abhijit Das}
\affiliation{
  \department{Department of Computer Science and Information Systems}
  \institution{Birla Institute of Technology and Science (BITS)}
  \city{Hyderabad}
  \country{India}
}
\email{abhijit.das@hyderabad.bits-pilani.ac.in}

\author{Yuankai Qi}
\affiliation{
  \department{School of Computing}
  \institution{Macquarie University}
  \city{Sydney}
  \country{Australia}
}
\email{yuankai.qi@mq.edu.au}

\author{Virendra Singh Shekhawat}
\affiliation{
  \department{Department of Computer Science and Information Systems}
  \institution{Birla Institute of Technology and Science (BITS)}
  \city{Pilani}
  \country{India}
}
\email{vsshekhawat@pilani.bits-pilani.ac.in}


\renewcommand{\shortauthors}{Nivedan \emph{et al.}\ }

\begin{abstract}
Vision-and-Language Navigation (VLN) is a multi-modal, cooperative task requiring agents to interpret human instructions, navigate 3D environments, and communicate effectively under ambiguity. This paper presents a comprehensive review of recent VLN advancements in robotics and outlines promising directions to improve multi-robot coordination. Despite progress, current models struggle with bidirectional communication, ambiguity resolution, and collaborative decision-making in the multi-agent systems. We review approximately 200 relevant articles to provide an in-depth understanding of the current landscape. Through this survey, we aim to provide a thorough resource that inspires further research at the intersection of VLN and robotics. We advocate that the future VLN systems should support proactive clarification, real-time feedback, and contextual reasoning through advanced natural language understanding (NLU) techniques. Additionally, decentralized decision-making frameworks with dynamic role assignment are essential for scalable, efficient multi-robot collaboration. These innovations can significantly enhance human-robot interaction (HRI) and enable real-world deployment in domains such as healthcare, logistics, and disaster response.
\end{abstract}



\received{20 February 2007}
\received[revised]{12 March 2009}
\received[accepted]{5 June 2009}

\maketitle

\section{INTRODUCTION}
\label{section_1}

VLN is a burgeoning field focused on creating embodied agents that both converse in natural language and autonomously navigate complex 3D environments \cite{anderson2018vision} \cite{chen2019touchdown}, \cite{thomason2020vision}, . In VLN tasks, agents interpret human instructions and leverage visual observations to traverse previously unseen spaces with increasing reliability. These agents integrate visual inputs and language instructions to generate navigation commands, provide verbal feedback, execute manipulation actions, and identify object locations\cite{wu2021visual} . As a hallmark of embodied AI (EAI) \cite{das2018embodied} , VLN has catalyzed extensions into vision-language-guided manipulation and outdoor navigation.

\begin{figure}[htbp]
    \centering
    \includegraphics[width=0.6\linewidth]{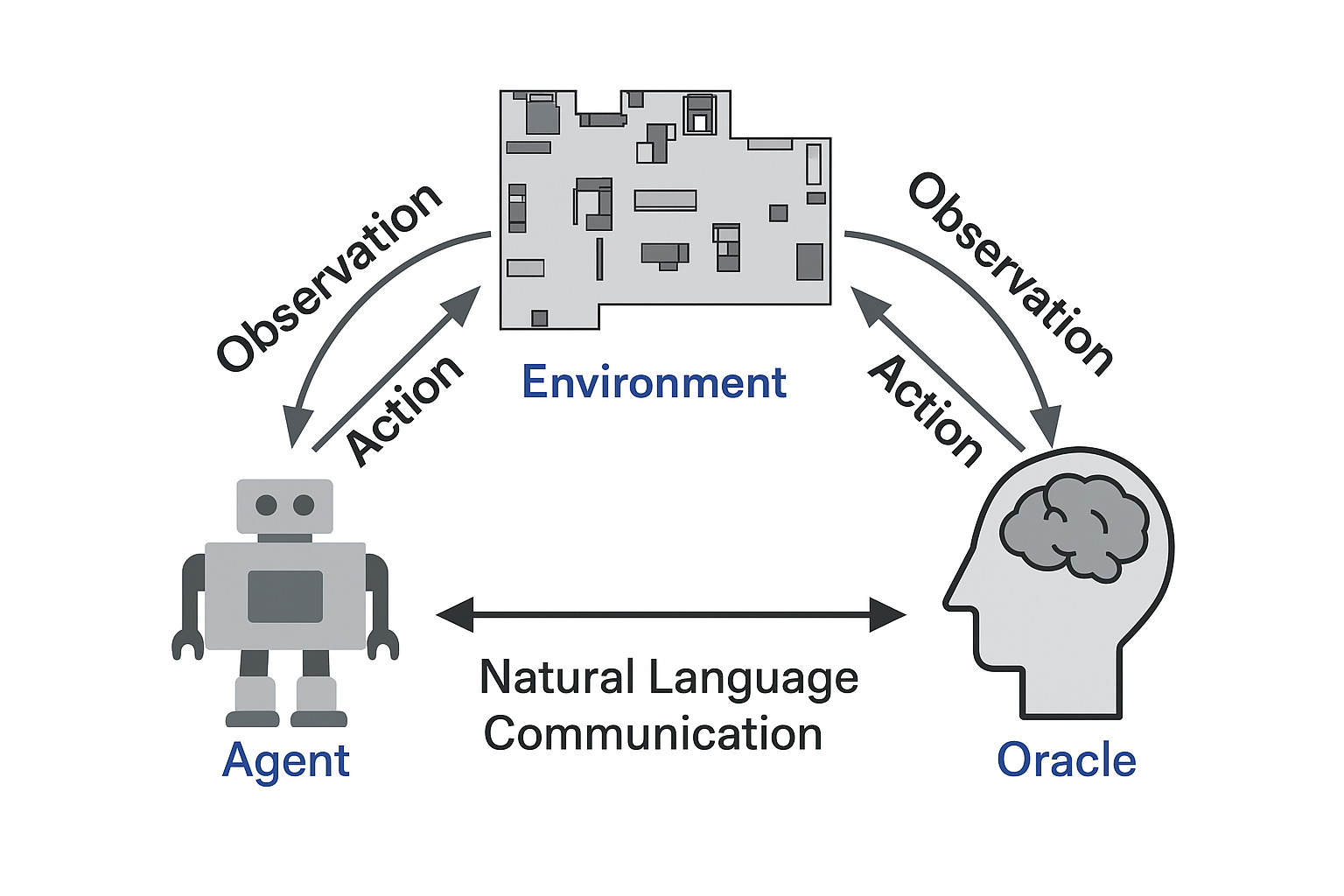}
    \Description{A schematic diagram showing agent-oracle communication in a 3D navigation environment. The agent receives language instructions and navigates accordingly.}
    \captionsetup{justification=justified}
    \caption{Schematic representation of VLN task as an interactive navigation via natural language\cite{gu2022vision} . The embodied agent perceives and acts within a 3D environment, while the oracle provides language-based guidance. Both the agent and oracle observe the environment and exchange information through natural language communication to achieve navigation objectives.}
    \label{fig1}
\end{figure}

VLN tasks center on an embodied agent and an oracle, often a human, operating in a 3D environment using natural language \cite{gu2022vision} , as illustrated in Figure~\ref{fig1}. The agent interprets instructions, requests clarifications when needed, and navigates or interacts with its surroundings. Meanwhile, the oracle observes progress and the environment state, providing guidance to ensure task success. By combining dialogue with visual navigation, VLN advances agents’ autonomy and adaptability in both simulated and real-world settings. VLN benchmarks have evolved along two key dimensions: communication and task scope. Some require agents to interpret a single instruction before navigating, while others support free-form dialogue with an oracle. Similarly, task objectives range from precise route following to dynamic exploration and object interaction. Even seemingly simple directives such as \textit{“Turn left, climb the stairs, enter the bathroom”} pose challenges for computational agents. They must decompose such instructions into sub-goals, ground each step in real-world objects and dynamics, recognize visual cues (e.g., identifying the bathroom), and execute actions accurately while knowing when to stop\cite{jain2019stay} .

Recent research underscores the importance of advancing multi-human and multi-robot systems to integrate robots seamlessly into society, calling for multidisciplinary efforts. Vision, in particular, enables robots to coordinate effectively with each other while naturally engaging human partners, which is a capability that VLN models are well-positioned to support. The existing literature on robotic vision for HRI and collaboration provides a strong foundation for VLN. Studies have examined robots with varying autonomy levels using vision for locomotion, manipulation, and visual communication in collaborative tasks. Furthermore, reviews of collaborative robotics emphasize safety, control performance, and intuitive interaction in industrial applications. Our paper examines existing works, challenges, and opportunities in developing robust collaborative VLN models that efficiently coordinate multiple robots while addressing the needs and preferences of human collaborators. The interdisciplinary nature of this field, as highlighted in\cite{buxbaum2020roadmap} , underscores the need for a structured approach to guide future research and development.

\subsection{Our Contributions}
\label{section_1.1}

While prior surveys on VLN have provided a strong foundation by categorizing tasks, methods, and challenges (Table ~\ref{Previous_works}), they largely adopt a Computer vision (CV) or AI-centric perspective, often overlooking the robotics-specific aspects of MRS and HRI. For example, Gu et al. \cite{gu2022vision} and Zhang et al. \cite{zhang2024vision} briefly mention collaborative elements in their future directions but do not explore them in depth, offering no dedicated analysis of decentralized decision-making or dynamic role assignment within MRS. Similarly, Park et al. \cite{park2023visual} and Wu et al. \cite{wu2021visual} concentrate on single-agent paradigms, missing the interdisciplinary bridge to HRI applications such as coordinated disaster response or logistics. Our survey directly addresses these gaps by presenting the first comprehensive resource at the intersection of CV, NLP and robotics, reviewing nearly more than 200 articles with particular emphasis on bidirectional communication, ambiguity resolution, and perception-driven collaboration in multi-agent contexts. We incorporate recent (post-2024) advancements in LLMs for contextual reasoning and proactive clarification, the areas partially covered in earlier works. By doing so, this survey aims to guide robotics research in real-world domains such as healthcare, outlining \textit{scalable MRS frameworks} and \textit{Sim-to-Real} innovations. In moving beyond the AI-generalized perspectives of prior surveys, we position our work as a timely and distinctive contribution to the field of embodied AI.

\usetikzlibrary{
    arrows.meta,
    positioning,
    calc,
    decorations.pathmorphing,
    bending
}

\definecolor{cIntro}{RGB}{210,70,70}
\definecolor{cEarly}{RGB}{230,130,60}
\definecolor{cMetrics}{RGB}{75,140,220}
\definecolor{cReview}{RGB}{70,170,130}
\definecolor{cMRS}{RGB}{150,90,200}
\definecolor{cLLM}{RGB}{240,180,60}
\definecolor{cAmb}{RGB}{120,120,120}
\definecolor{cFuture}{RGB}{250,160,40}

\tikzset{
    mainbox/.style={
        draw,
        thick,
        rounded corners=5pt,
        fill=#1!20,
        font=\large,
        inner sep=5pt,
        minimum width=3cm,
        align=center
    },
    subbox/.style={
        draw,
        thick,
        rounded corners=4pt,
        fill=#1!15,
        font=\large,
        inner sep=4pt,
        minimum width=2.8cm,
        align=center
    },
    arc/.style={
        thick,#1!80,
        -{Latex[bend]},
        bend left=20
    }
}

\begin{figure*}[t]
\centering
\resizebox{\textwidth}{!}{
\begin{tikzpicture}[
    node distance=0.9cm,
    every node/.style={align=center}
]


\node[mainbox=cIntro] (intro) {Introduction\\ (§\ref{section_1})};
\node[mainbox=cEarly, right=0.5cm of intro] (early) {Early\\Foundations\\(§\ref{section_2})};
\node[mainbox=cMetrics, right=0.5cm of early] (metrics) {VLN Evaluation\\Metrics\\ (§\ref{section_3})};
\node[mainbox=cReview, right=0.5cm of metrics] (review) {Review of recent\\Developments\\(§\ref{section_4})};
\node[mainbox=cMRS, right=0.5cm of review] (mrs) {VLN-based \\MRS \& HRI\\(§\ref{section_5})};
\node[mainbox=cLLM, right=0.5cm of mrs] (llm) {LLMs \& Reasoning \\in VLN\\(§\ref{section_6})};
\node[mainbox=cAmb, right=0.5cm of llm] (amb) {Ambiguity in \\Language Instructions\\(§\ref{sec:ambiguity})};
\node[mainbox=cFuture, right=0.5cm of amb] (future) {Conclusion \& \\Future Directions\\(§\ref{Future})};


\node[subbox=cIntro, below=0.6cm of intro] (intro1)
    {Our Contributions\\ (§\ref{section_1.1})};

\node[subbox=cEarly, below=0.55cm of early] (early1)
    {VLN Datasets\\ Simulators};

\node[subbox=cReview, below=0.55cm of review] (rev1)
    {Initial Work\\ (§\ref{section_4.1})};
\node[subbox=cReview, below=0.4cm of rev1] (rev2)
    {Diversification\\ (§\ref{section_4.2})};
\node[subbox=cReview, below=0.4cm of rev2] (rev3)
    {Current Breakthroughs\\ (§\ref{section_4.3})};

\node[subbox=cFuture, below=0.55cm of future] (f1)
    {Limited HRI\\Capabilities\\(§\ref{future_1})};
\node[subbox=cFuture, below=0.4cm of f1] (f2)
    {Ambiguous Instructions\\(§\ref{future_2})};
\node[subbox=cFuture, below=0.4cm of f2] (f3)
    {Lack of Robust\\Multi-Robot Coordination\\(§\ref{future_3})};
\node[subbox=cFuture, below=0.4cm of f3] (f4)
    {Sim2Real Transfer\\Challenge\\(§\ref{future_4})};


\draw[arc=cIntro] (intro) to (early);
\draw[arc=cEarly] (early) to (metrics);
\draw[arc=cMetrics] (metrics) to (review);

\draw[arc=cReview] (review) to (mrs);
\draw[arc=cMRS] (mrs) to (llm);

\draw[arc=cLLM] (llm) to (amb);
\draw[arc=cAmb] (amb) to (future);


\draw[arc=cIntro] (intro) -- (intro1);
\draw[arc=cEarly] (early) -- (early1);
\draw[arc=cReview] (review) -- (rev1);
\draw[arc=cReview] (rev1) -- (rev2);
\draw[arc=cReview] (rev2) -- (rev3);

\draw[arc=cFuture] (future) -- (f1);
\draw[arc=cFuture] (f1) -- (f2);
\draw[arc=cFuture] (f2) -- (f3);
\draw[arc=cFuture] (f3) -- (f4);


\end{tikzpicture}
}
\caption{Overview of this survey.}
\Description{This figure shows the overview of the paper}
\label{fig:overview}
\end{figure*}
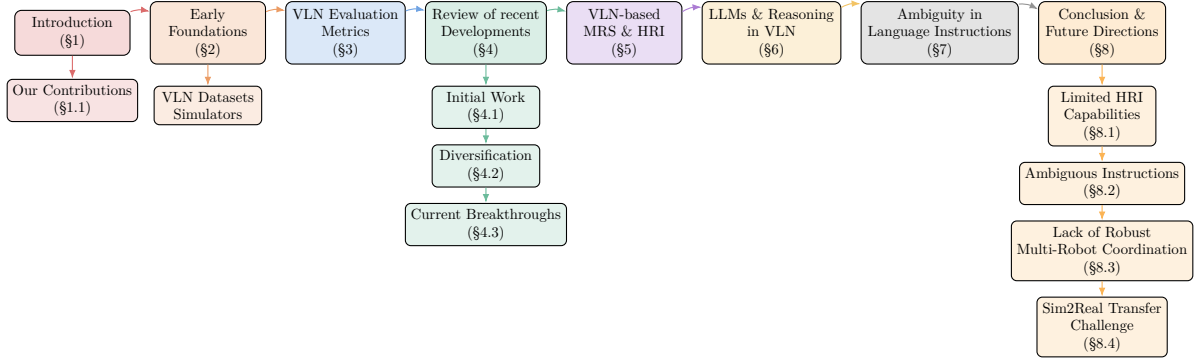

\begin{figure}[htbp]
    \centering
    \Description{Datasets and Environments for Visual Navigation.}
    \includegraphics[width=0.7\textwidth]{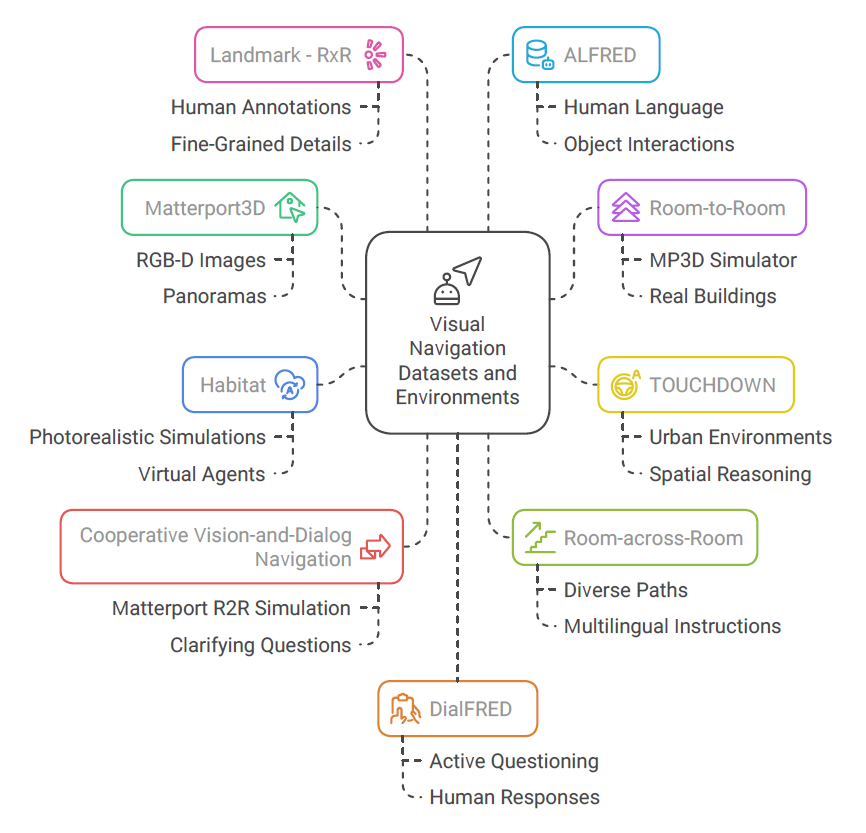}
    \caption{\small Prominent VLN Datasets and Environments for Visual Navigation.}
    \label{fig3}
\end{figure}

The comparisons of our paper with the previous survey papers like \cite{gu2022vision}, \cite{park2023visual}, \cite{wu2021visual}, and \cite{zhang2024vision} have been tabulated in the Table ~\ref{Previous_works}. Fig. \ref{fig:overview} shows the structure of this survey. The paper is divided into eight sections as follows: Section ~\ref{section_2} emphasizes the early VLN research foundations, datasets (Fig. \ref{fig3}), and simulators (Table ~\ref{datasets}). Section ~\ref{section_3} discusses the various VLN evaluation metrics. The recent developments and important state-of-the-art (SoTA) research works are discussed in Section ~\ref{section_4}, sub-divided as initial work, diversification and the latest contributions. Section ~\ref{section_5}, discusses the crux of this paper, \textbf{VLN based MRS and HRI}. The section ~\ref{section_6} talks about the very recent applications of LLMs and reasoning in VLN and section ~\ref{sec:ambiguity} elaborates the existing works which considers the ambiguity present in the natural instructions. Based on these, the paper concludes with a few remarks on the future research directions in the field of VLN for MRS in section ~\ref{Future}.

\begin{longtable}{|p{2.3cm}|p{1.5cm}|p{3.5cm}|p{3.5cm}|p{1.6cm}|p{1.6cm}|}
\caption{Common notion of comparison with the previous surveys.}
\label{Previous_works} \\
\hline
\rowcolor[HTML]{C0C0C0} 
\small \textbf{Survey paper, Year} & 
\small \textbf{Venue, \# Citations} & 
\small \textbf{Primary contributions} & 
\small \textbf{Future work proposed} & 
\small \textbf{Surveyed MRS \& HRI in VLN?} & 
\small \textbf{Surveyed LLMs \& reasoning in VLN?} \\ 
\hline
\endfirsthead

\multicolumn{6}{c}{{\small \tablename\ \thetable{} -- Continued from previous page}} \\ \hline
\rowcolor[HTML]{C0C0C0} 
\small \textbf{Survey paper, Year} & 
\small \textbf{Venue, \# Citations} & 
\small \textbf{Primary contributions} & 
\small \textbf{Future work proposed} & 
\small \textbf{Surveyed MRS \& HRI in VLN?} & 
\small \textbf{Surveyed LLMs \& reasoning in VLN?} \\ 
\hline
\endhead

\hline \multicolumn{6}{|r|}{{\small Continued on next page}} \\ \hline
\endfoot

\endlastfoot

\small Vision-and-Language Navigation: A Survey of Tasks, Methods, and Future Directions \cite{gu2022vision} (2022) &
\small ACL Annual Meeting (\textbf{161 citations}) &
\small Discusses VLN agents as societal entities, analyzing how their tasks vary by \textbf{communication level} vs. \textbf{task objective}, and how agents can be evaluated. &
\small Collaborative VLN between multiple agents or between humans and agents; exploring diverse cultural and linguistic environments. &
\small No & \small No \\ 
\hline

\small Visual Language Navigation: A Survey and Open Challenges \cite{park2023visual} (2023) &
\small Artificial Intelligence Review, Vol. 56(1) (\textbf{48 citations}) &
\small Defines a taxonomy for VLN techniques and analyzes them through four lenses: \textbf{representation learning}, \textbf{reinforcement learning}, \textbf{components}, and \textbf{evaluation}. &
\small To enhance human-agent interaction; adopt video-based imitation learning and multimodal synchronization. &
\small No & \small No \\ 
\hline

\small Vision-Language Navigation: A Survey and Taxonomy \cite{wu2021visual} (2024) &
\small Neural Computing and Applications, Vol. 36 (\textbf{40 citations}) &
\small Categorizes navigation tasks by instruction frequency into \textbf{single-turn} (goal- or route-oriented) and \textbf{multi-turn} (passive or interactive). &
\small Incorporating knowledge bases (e.g., common-sense or algorithmic) and memory for real-world deployment and improved reasoning. &
\small No & \small No \\ 
\hline

\small Vision-and-Language Navigation Today and Tomorrow: A Survey in the Era of Foundation Models \cite{zhang2024vision} (2024) &
\small Transactions on Machine Learning Research (\textbf{37 citations}) &
\small Presents a top-down review of embodied planning and reasoning; identifies directions like \textbf{world modeling}, \textbf{human modeling}, and \textbf{agent design} for grounded reasoning. &
\small Calls for expanding VLN datasets, enhancing 3D representations, and improving commonsense transfer for embodied agents. &
\small No & \small Yes (partially) \\ 
\hline

\small \textbf{Ours} &
\small -- &
\small Reviews VLN tasks, datasets, and simulators, emphasizing \textbf{MRS}, \textbf{HRI}, and \textbf{LLMs} for reasoning, decision-making, and collaborative navigation. &
\small Proposes proactive clarification, real-time feedback, and contextual reasoning through advanced NLU. Recommends realistic simulation environments (Section~\ref{Future}). &
\small \textbf{Yes} & \small \textbf{Yes} \\ 
\hline
\end{longtable}

\section{Early Foundations}
\label{section_2}

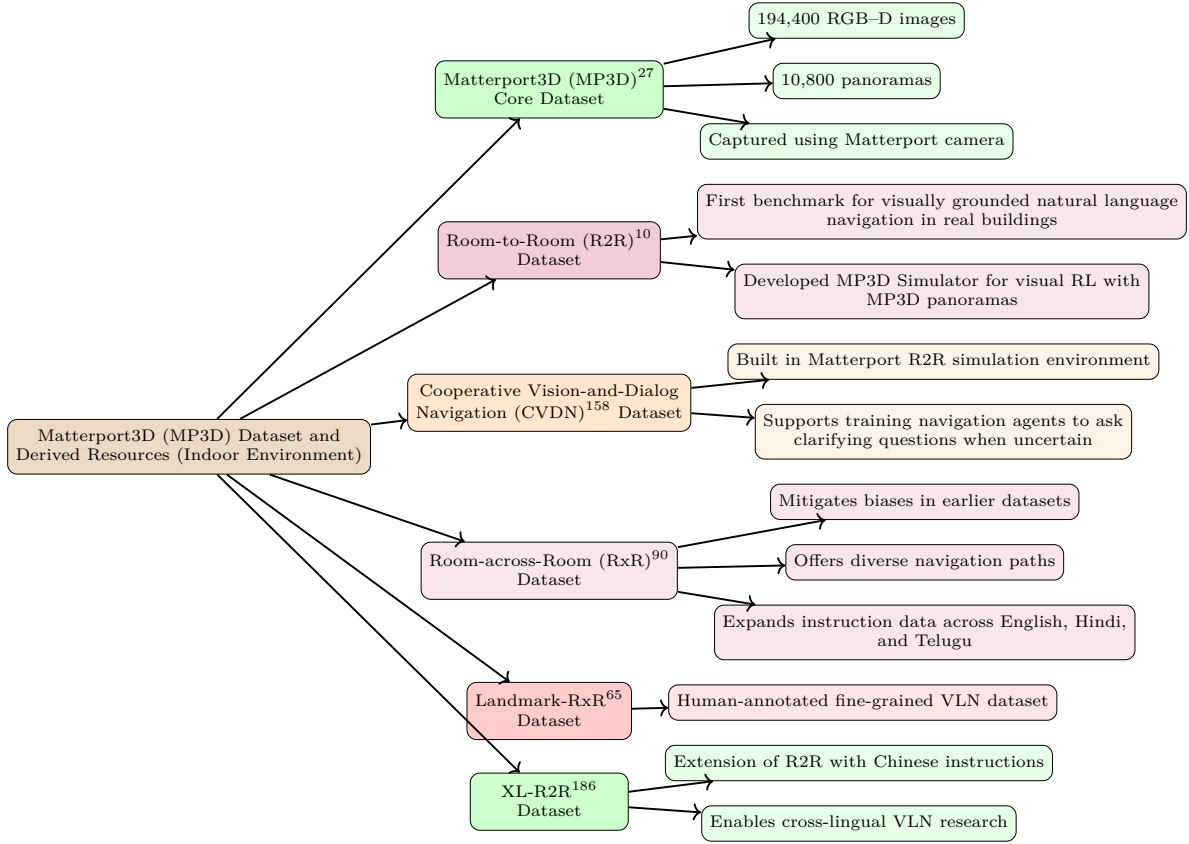
\begin{figure}[htbp]
\centering
\Description{Hierarchical structure of the Matterport3D (MP3D) dataset and its derived VLN datasets such as R2R, RxR, XL-R2R, Landmark-RxR, and CVDN.}
\adjustbox{max width=\textwidth}{
\begin{forest}
for tree={
    grow'=east,
    draw,
    rounded corners,
    align=center,
    edge={->, thick},
    s sep=10pt,
    l sep=15pt,
    minimum width=2.3cm,
    font=\scriptsize,
}
[{Matterport3D (MP3D) Dataset and \\Derived Resources (Indoor Environment)}, fill=brown!30
  [{Matterport3D (MP3D)\cite{chang2017matterport3d}\\ Core Dataset}, fill=green!20
    [{194{,}400 RGB\textendash D images}, fill=green!10]
    [{10{,}800 panoramas}, fill=green!10]
    [{Captured using Matterport camera}, fill=green!10]
  ]
  [{Room-to-Room (R2R)\cite{anderson2018vision} \\Dataset}, fill=purple!20
    [{First benchmark for visually grounded natural language \\navigation in real buildings}, fill=purple!10]
    [{Developed MP3D Simulator for visual RL with \\MP3D panoramas}, fill=purple!10]
  ]
    [{Cooperative Vision-and-Dialog \\Navigation (CVDN)\cite{thomason2020vision} Dataset}, fill=orange!20
    [{Built in Matterport R2R simulation environment}, fill=orange!10]
    [{Supports training navigation agents to ask \\clarifying questions when uncertain}, fill=orange!10]
  ]
  [{Room-across-Room (RxR)\cite{ku2020room} \\Dataset}, fill=purple!10
    [{Mitigates biases in earlier datasets}, fill=purple!10]
    [{Offers diverse navigation paths}, fill=purple!10]
    [{Expands instruction data across English, Hindi, \\and Telugu}, fill=purple!10]
  ]
    [{Landmark-RxR\cite{he2021landmark} \\Dataset}, fill=red!20
    [{Human-annotated fine-grained VLN dataset}, fill=red!10]
  ]
  [{XL-R2R\cite{yan2019cross} \\Dataset}, fill=green!20
    [{Extension of R2R with Chinese instructions}, fill=green!10]
    [{Enables cross-lingual VLN research}, fill=green!10]
  ]
]
\end{forest}
}
\caption{Hierarchical structure of the Matterport3D (MP3D) dataset and its derived VLN resources.}
\label{fig:mp3d-tree}
\end{figure}

Over time, VLN has been extensively studied in both photorealistic simulators and real-world environments, resulting in various benchmarks with differing problem formulations. In VLN tasks, datasets offer visual assets and scenes, while simulators render these assets, providing an environment for the VLN agent. We have categorized them as MP3D dataset and its derived VLN resources (Indoor Environment) (Fig. \ref{fig:mp3d-tree}), VLN outdoor datasets (Fig. \ref{fig:oudoor-tree}), embodied AI and other practical VLN datasets (Fig. \ref{fig:pratical-tree}) and Simulation Environments and Platforms for VLN (Fig. \ref{fig:simulators-tree}). 

Initially Chang \emph{et al.}\ \cite{chang2017matterport3d} introduced the \textbf{Matterport3D (MP3D)} dataset, which offered new research opportunities for learning about indoor home environments. This dataset includes 194,400 RGB-D images captured in 10,800 panoramas using a Matterport camera. Anderson \emph{et al.} \cite{anderson2018vision} presented the \textbf{Room-to-Room (R2R)} dataset, the first benchmark for visually grounded natural language navigation in real buildings. They also developed the \textbf{MP3D Simulator}, a framework for visual reinforcement learning (RL) using the MP3D panoramic dataset, and established several baselines by applying sequence-to-sequence \cite{bahdanau2014neural} neural networks.  

Chen \emph{et al.} \cite{chen2019touchdown} introduced \textbf{TOUCHDOWN}, a dataset designed for natural language navigation and spatial reasoning in real-world urban environments. Savva \emph{et al.}\ \cite{savva2019habitat} introduced \textbf{Habitat}, a high-performance platform for EAI research, enabling the training of virtual agents in photorealistic 3D simulations. Habitat comprises two main components: \textbf{Habitat-Sim}, a fast and flexible 3D simulator capable of rendering thousands of frames per second, and \textbf{Habitat-API}, a modular library supporting the development of EAI tasks such as navigation, instruction following, and question answering. The \textbf{Cooperative Vision-and-Dialog Navigation (CVDN)}  \cite{thomason2020vision} dataset, developed in the Matterport R2R simulation environment, supports training navigation agents that can ask clarifying questions when uncertain. \textit{Navigation from Dialog History (NDH)} is a task featuring over 7,000 annotated instances, and evaluates a seq-2-seq model that predicts navigation actions from dialog history. Their work points to promising directions, such as cooperative multi-agent learning and more advanced models to bridge the gap toward human-level navigation performance.



\textbf{Room-across-Room (RxR)} \cite{ku2020room}, another VLN dataset that mitigates biases present in earlier datasets by offering a greater diversity of paths and significantly expanding instruction data across three languages, \textbf{English, Hindi, and Telugu}. Cross-modal alignment, ensuring predicted trajectories precisely follow instructions, remains a key VLN challenge. To address this, He \emph{et al.}\ introduce \textbf{Landmark‑RxR} \cite{he2021landmark}, a human‑annotated fine‑grained VLN dataset. \textbf{ALFRED} \cite{shridhar2020alfred} is a benchmark designed to link human language with actions, behaviors, and object interactions in interactive visual environments. Unlike traditional tasks, agents in ALFRED must complete complex tasks specified through natural language, involving both navigation and object manipulation. \textbf{DialFRED} \cite{gao2022dialfred}, an extension of the ALFRED benchmark that enables agents to actively ask questions and leverage human responses for better task execution. DialFRED addresses two key challenges: resolving language ambiguities through clarification and planning long-horizon action sequences while recovering from failures. The benchmark is publicly released to foster innovation across the robotics and EAI communities. 

\textbf{ReALFRED} \cite{kim2024realfred} benchmark extends ALFRED by incorporating real-world scenes, objects, and layouts, enabling agents to follow natural language instructions in larger, 3D-captured, multi-room environments. Evaluation of existing ALFRED methods on ReALFRED reveals consistent performance drops, highlighting the need for approaches better suited to realistic settings, and also, current systems support only English, limiting accessibility for users speaking other languages. Future research could explore (1) incorporating more complex bimanual tasks and (2) enabling multilingual interfaces to accommodate diverse user populations. In that direction, the \textbf{XL-R2R} \cite{yan2019cross} dataset extends the R2R benchmark with Chinese instructions, enabling cross-lingual VLN research. They explore zero-shot navigation, training on English while testing in Chinese, and find the model performs competitively even without target-language data.

The \textbf{Remote Embodied Visual Referring Expressions in Real 3D Indoor Environments (REVERIE)} \cite{qi2020reverie} dataset introduces a more practical VLN task. Unlike earlier tasks focused solely on action sequences or answers, REVERIE requires agents to predict bounding boxes for target objects. The dataset includes 10,567 panoramas across 90 buildings, 4,140 target objects, and 21,702 crowd-sourced instructions averaging 18 words each. A major challenge in VLN is grounding multilingual instructions and navigating unseen environments. \textbf{CLEAR} \cite{li2022clear} (Cross-Lingual and Environment-Agnostic Representations), where an agent learns shared, visually aligned language representations across English, Hindi, and Telugu using visually grounded text pairs was proposed. Additionally, the agent develops an environment-agnostic visual representation by aligning semantically similar images across different settings, reducing bias from low-level visual cues. 

\begin{figure}[htbp]
\centering
\Description{Hierarchical structure of the outdoor dataset.}
\adjustbox{max width=\textwidth}{
\begin{forest}
for tree={
    grow'=east,
    draw,
    rounded corners,
    align=center,
    edge={->, thick},
    s sep=10pt,
    l sep=15pt,
    minimum width=2.3cm,
    font=\scriptsize,
}
[{Natural Language Navigation Datasets \\for Urban and Outdoor Environments}, fill=brown!30
  [{TOUCHDOWN\cite{chen2019touchdown} \\Dataset}, fill=green!20
    [{Designed for natural language navigation and \\spatial reasoning}, fill=green!10]
    [{Real-world urban environments}, fill=green!10]
  ]
  [{StreetNav\cite{hermann2020learning} \\Dataset}, fill=purple!20
    [{Instruction following task using \\Google Street View}, fill=purple!10]
    [{Combines simulated control with real-world \\ambiguity}, fill=purple!10]
  ]
    [{StreetLearn\cite{mirowski2019streetlearn} \\Dataset}, fill=orange!20
    [{First-person, partially observed environment on \\Google Street View imagery}, fill=orange!10]
    [{Supports end-to-end learning and goal-driven \\navigation baselines}, fill=orange!10]
  ]
  [{Talk2Nav\cite{vasudevan2021talk2nav} \\Dataset}, fill=purple!10
    [{Large-scale dataset with \\10{,}714 routes}, fill=purple!10]
    [{Crowd-sourced verbal directions in Google \\Street View environment}, fill=purple!10]
    [{Annotated with mined landmark anchors and \\local way-finding cues}, fill=purple!10]
  ]
]
\end{forest}
}
\caption{Hierarchical structure of the prominent VLN outdoor datasets.}
\label{fig:oudoor-tree}
\end{figure}
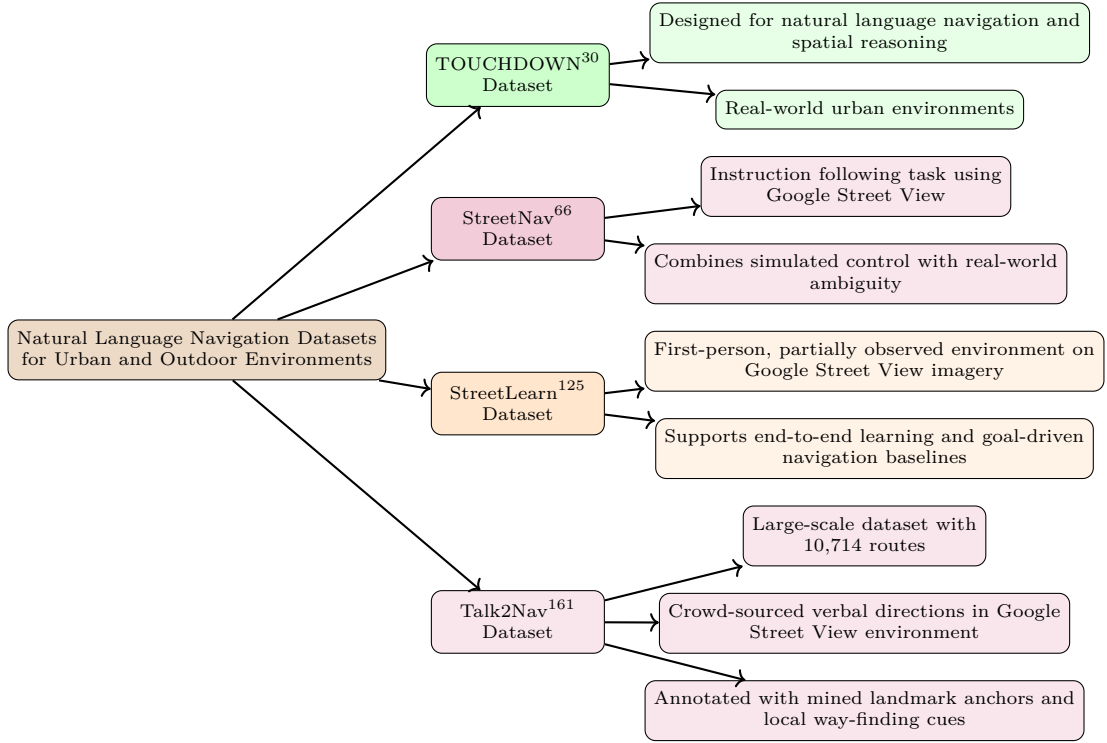
\begin{figure}[htbp]
\centering
\Description{Hierarchical structure of the pratical datasets.}
\adjustbox{max width=\textwidth}{
\begin{forest}
for tree={
    grow'=east,
    draw,
    rounded corners,
    align=center,
    edge={->, thick},
    s sep=10pt,
    l sep=15pt,
    minimum width=2.3cm,
    font=\scriptsize,
}
[{Other VLN Datasets}, fill=brown!30
  [{Embodied AI and Visual Navigation \\Benchmarks and Extensions}, fill=green!20
    [{ALFRED\cite{shridhar2020alfred} \\Benchmark}, fill=green!10
        [{Links human language with action, \\behavior, and object interaction}, fill=green!5]
        [{Interactive visual environments}, fill=green!5]
    ]
    [{DialFRED\cite{gao2022dialfred}}, fill=green!10
        [{Extension of ALFRED}, fill=green!5]
        [{Enables agents to ask questions and \\use human responses for better task execution}, fill=green!5]
    ]
    [{ReALFRED\cite{kim2024realfred} \\Benchmark}, fill=green!10
        [{Extends ALFRED with real-world \\scenes, objects, layouts}, fill=green!5]
        [{Supports navigation in larger, 3D-captured \\multi-room environments}, fill=green!5]
    ]
  ]
  [{Practical VLN Tasks \\and Datasets}, fill=purple!20
    [{REVERIE\cite{qi2020reverie}}, fill=purple!10
    [{10{,}567 panoramas across \\90 buildings}, fill=purple!5]
    [{4{,}140 target objects}, fill=purple!5]
    [{21{,}702 crowd-sourced \\instructions averaging 18 words}, fill=purple!5]
    [{Introduces a more practical \\VLN task}, fill=purple!5]
    ]
  ]
]
\end{forest}
}
\caption{Hierarchical structure of the Embodied AI and other practical VLN datasets.}
\label{fig:pratical-tree}
\end{figure}
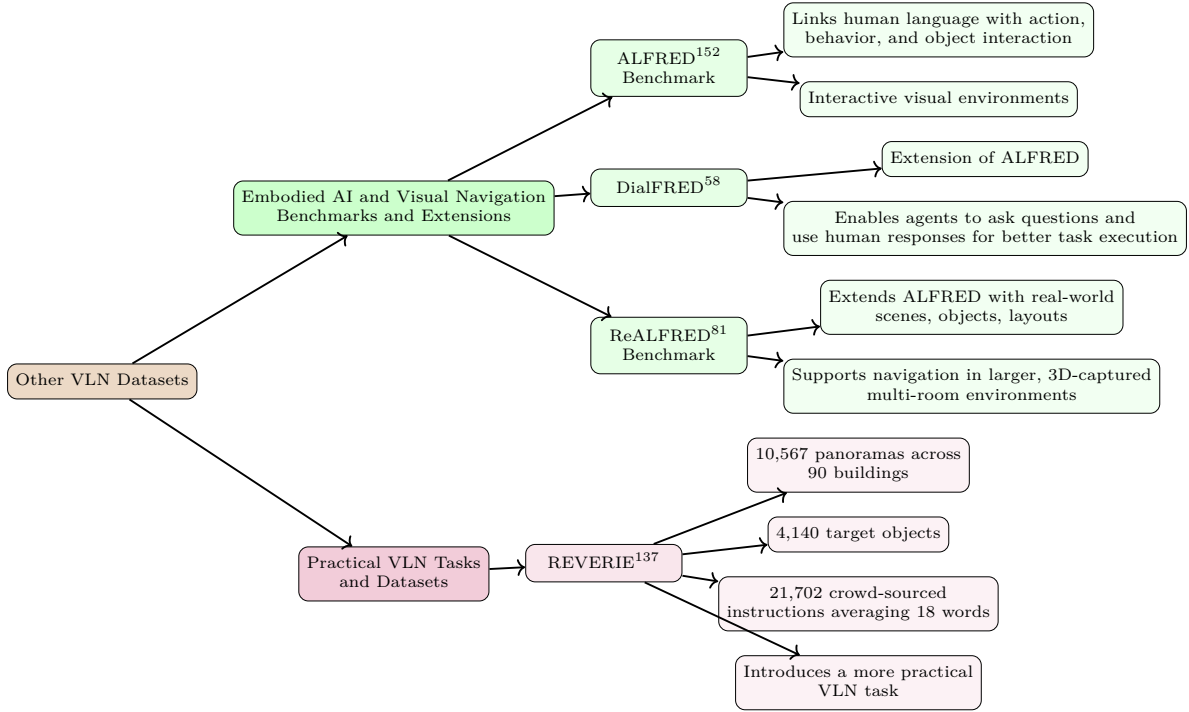


Navigation drives advances in perception, planning, memory, exploration, and optimization. Yet most benchmarks rely on static datasets, such as recorded trajectories, that do not support interactive decision-making or RL. To address this gap, \textbf{StreetLearn} \cite{mirowski2019streetlearn} offers a first‑person, partially observed environment built on Google Street View\footnotemark imagery, enabling end‑to‑end learning and providing baselines for goal‑driven navigation tasks. Real-world navigation challenges spur advances in language grounding, planning, and vision. Considering that \textbf{StreetNav} \cite{hermann2020learning}, an instruction following task built on Google Street View that blends simulated control with real-world ambiguity. Agents learn to interpret driving directions in visually accurate, multi‑city environments. By enforcing a strict train/test split across unseen cities, StreetNav rigorously evaluates an agent’s generalization, mirroring the human ability to navigate new locales. 
\footnotetext{\url{https://developers.google.com/maps/documentation/streetview/overview}}

Robots’ growing societal role demands intuitive human–robot communication, particularly for verbal navigation. Vasudevan et al. introduce \textbf{Talk2Nav} \cite{vasudevan2021talk2nav}, a large-scale dataset of 10,714 routes with crowd-sourced verbal directions in a Google Street View-based environment, annotated via mined landmark anchors and local wayfinding cues. Building on navigation efficiency, Ke \emph{et al.}\ \cite{ke2019tactical} introduced the \textbf{FAST} Navigator (Frontier Aware Search with backtracking), a general framework for action decoding. Their approach enables agents to navigate from source to target locations in unseen environments by acting greedily while leveraging global signals to backtrack when needed, improving navigation performance.

Generalizing VLN agents to unseen 3D environments requires robustness to both low‑level (color, texture) and high‑level (layout) variations, typically addressed via multi‑level data augmentation. In that regard, Wu \emph{et al.}\ introduce \textbf{House3D} \cite{wu2018building}, an extensible environment of 45,622 richly annotated \textbf{SUNCG} \cite{song2017semantic} scenes from studios to multi‑story homes, that supports scene, pixel, and task‑level augmentations. Whereas Fried \emph{et al.}\ \cite{fried2018speaker} proposed an embedded speaker model for VLN that synthesizes new instructions for data augmentation and enables pragmatic reasoning by assessing how well candidate action sequences align with a given instruction. 

\begin{figure}[htbp]
\centering
\Description{Hierarchical structure of the Simulation Environments and Platforms for VLN research.}
\adjustbox{max width=\textwidth}{
\begin{forest}
for tree={
    grow'=east,
    draw,
    rounded corners,
    align=center,
    edge={->, thick},
    s sep=10pt,
    l sep=15pt,
    minimum width=2.3cm,
    font=\scriptsize,
}
[{Simulation Environments and \\Platforms for Embodied AI}, fill=brown!30
  [{Habitat\cite{savva2019habitat} \\Platform}, fill=green!20
    [{High-performance platform for \\Embodied AI research}, fill=green!10]
    [{Enables training virtual agents in \\photorealistic 3D simulations}, fill=green!10]
  ]
  [{House3D\cite{wu2018building} \\Environment}, fill=purple!20
    [{Extensible environment with 45{,}622 richly \\annotated SUNCG scenes}, fill=purple!10]
    [{Supports scene, pixel, and task-level \\augmentations}, fill=purple!10]
  ]
    [{Gibson\cite{xia2018gibson} \\Environment}, fill=orange!20
    [{Simulation platform for developing real-world \\perception in active agents}, fill=orange!10]
  ]
  [{iGibson 1.0\cite{shen2020lp}}, fill=purple!10
    [{Simulation environment for robotic solutions in \\large-scale realistic settings}, fill=purple!10]
    [{15 fully interactive home-sized scenes \\with 108 rooms}, fill=purple!10]
    [{Includes rigid and articulated \\objects replicating real-world layouts}, fill=purple!10]
  ]
    [{AI2-THOR\cite{kolve2017ai2} \\Framework}, fill=red!20
    [{Photo-realistic 3D simulation for \\visual AI research}, fill=red!10]
    [{Features interactive indoor environments \\for navigation and object manipulation}, fill=red!10]
  ]
  [{HANNA\cite{nguyen2019help} \\Simulator ("Help, Anna!")}, fill=green!20
    [{Photorealistic simulator where agents request and \\interpret natural language and visual cues}, fill=green!10]
    [{Simulated assistants (ANNA) help locate objects \\via dialogue}, fill=green!10]
  ]
]
\end{forest}
}
\caption{Hierarchical structure of the Simulation Environments and Platforms for VLN research.}
\label{fig:simulators-tree}
\end{figure}
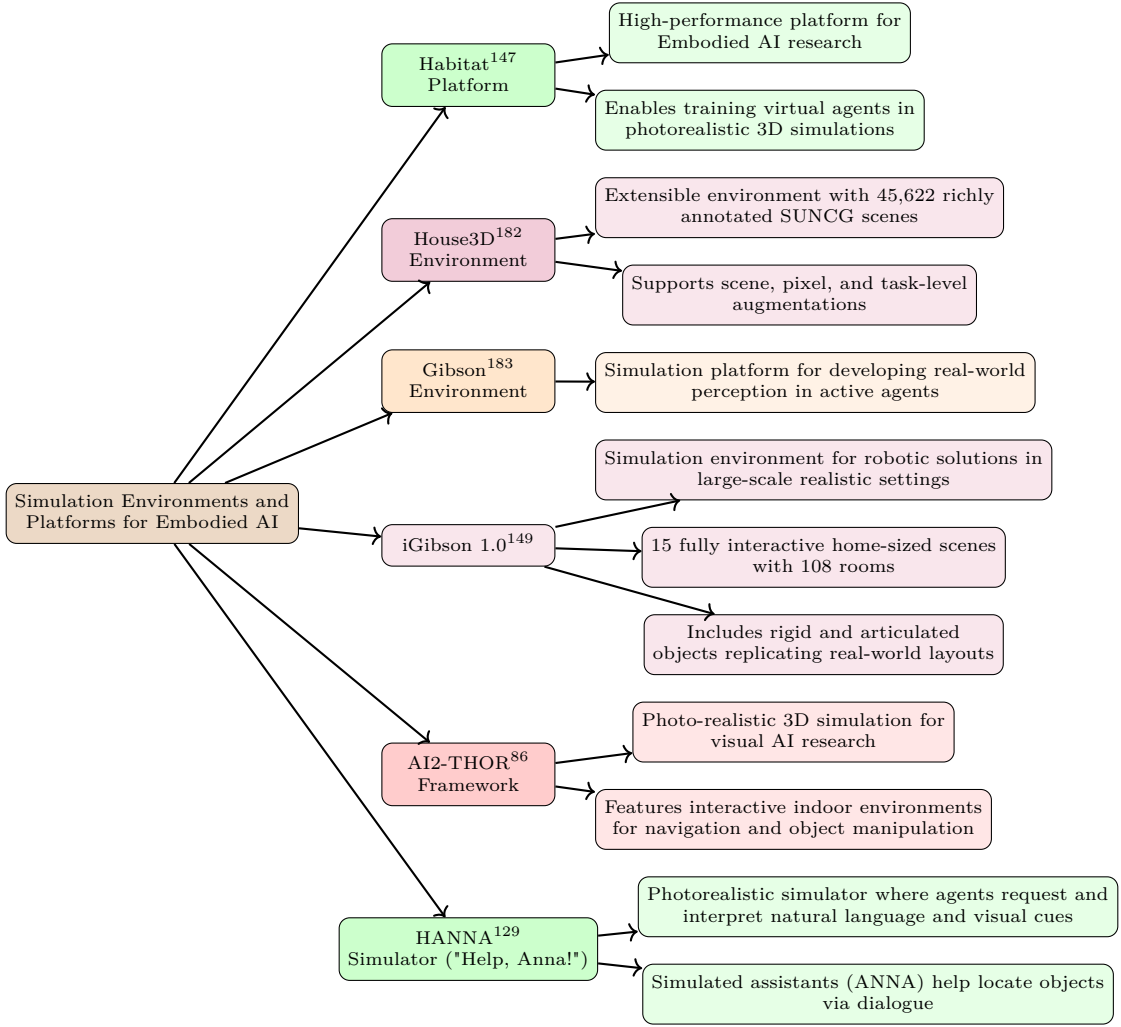

Training visual perception in the physical world is challenging due to high costs, slow learning, and robot fragility. To address this, \textbf{Gibson Environment} \cite{xia2018gibson}, a simulation platform was designed to develop real-world perception in active agents. Unlike synthetic environments, Gibson virtualizes real spaces, featuring over 1,400 floor plans from 572 buildings, and supports physically embodied agents. Key features include (i) real-world semantic complexity, (ii) an internal “Goggles” system for real-world deployment without domain adaptation, and (iii) physics-based constraints to enhance realism. Whereas \textbf{iGibson 1.0} \cite{shen2020lp} is a simulation environment designed to advance robotic solutions for interactive tasks in large-scale, realistic settings. It features 15 fully interactive, home-sized scenes with 108 rooms, populated by both rigid and articulated objects, replicating real-world layouts. iGibson also offers a human interface, allowing users to navigate and interact with objects, such as pulling, pushing, and placing, through simple mouse and keyboard commands, enhancing accessibility for research and development.

\textbf{ViZDoom} \cite{kempka2016vizdoom} is a lightweight and customizable platform for vision-based RL in semi-realistic 3D environments from a first-person perspective. Unlike 2D Atari games, ViZDoom offers a more realistic test-bed by leveraging the classic game Doom, allowing agents to learn from raw pixel input. The bots achieved human-like performance, highlighting ViZDoom’s potential for advancing visual RL in immersive environments. \textbf{AI2-THOR} \cite{kolve2017ai2} is a photo-realistic 3D simulation framework for visual AI research. It features interactive indoor environments where agents can navigate and manipulate objects to complete tasks. It supports a wide range of research areas, including deep reinforcement learning (DRL), imitation learning (IL), planning, visual question answering, and cognitive modeling. Its primary aim is to advance the development of visually intelligent agents through interactive learning in realistic settings. Whereas \textbf{HANNA} \cite{nguyen2019help} (“Help, Anna!”) is a photorealistic simulator where agents request and interpret natural language and visual cues from simulated assistants (ANNA) to locate objects. A memory-augmented neural agent with hierarchical decision levels and an IL algorithm that avoids past errors while estimating future progress was proposed.

Recently, foundation models \cite{bommasani2021opportunities} ranging from early pre-trained models like \textbf{Bidirectional Encoder Representations from Transformers (BERT)} \cite{devlin2018bert} to modern \textbf{Large Language Models (LLMs)} \cite{touvron2023llama} and \textbf{Vision-Language Models (VLMs)} \cite{radford2021learning} have demonstrated remarkable capabilities in multi-modal understanding, reasoning, and cross-domain generalization \cite{zhang2024vision}. Utilizing the LLMs and the VLMs for language understanding, visual perception, cross-model modeling, planning, and decision-making in language-guided navigation and manipulation is an emerging direction in EAI \cite{ahn2022can}, \cite{shah2023lm}. Further, Zero-shot approaches leverage pretrained VLMs and LLMs to enable agents to utilize prior knowledge for decision-making. These models exhibit strong performance across vision and language tasks without task-specific training \cite{sun2024survey}. However, their learned semantic and spatial knowledge remains underutilized in the Object Navigation (ObjectNav) task, presenting a valuable direction for future exploration.

\begin{table}[htbp]
\centering
\small
\caption{Comparisons of important VLN datasets.}
\label{datasets}
\begin{tabularx}{\textwidth}{|>{\justify\arraybackslash}p{2.3cm}
                             |>{\justify\arraybackslash}p{5.4cm}
                             |>{\justify\arraybackslash}p{4cm}
                             |>{\justify\arraybackslash}p{2.7cm}|}
\hline
\rowcolor[HTML]{C0C0C0} \textbf{Dataset} & \textbf{Purpose} & \textbf{Dataset Size} & \textbf{Simulator} \\ 
\hline
\textbf{R2R} \cite{anderson2018vision} and \textbf{Room-for-Room (R4R)} \cite{jain2019stay} & R2R is a first benchmark dataset for visually-grounded natural language navigation in real buildings. R4R is an algorithmically produced extension of R2R, which includes larger and more diverse reference paths. & Contains \textbf{21,567} open vocabulary, crowd-sourced navigation instructions with an average length of \textbf{29} words. & Matterport3D. \\ 
\hline
\textbf{TOUCHDOWN} \cite{chen2019touchdown} & TOUCHDOWN task and dataset: an agent must first follow navigation instructions in a real-life visual urban environment, and then identify a location described in natural language to find a hidden object at the goal position. & The data contains \textbf{9,326} examples of English instructions and spatial descriptions paired with demonstrations. The environment includes 29,641 panoramas and 61,319 edges from New York City. & Google Street View \\ 
\hline
\textbf{REVERIE} \cite{qi2020reverie} & Given a natural language instruction that represents a practical task to perform, an agent must navigate and identify a remote object in real indoor environments. The REVERIE task differs from previous works that only output a simple answer or a series of actions, as they ask the agent to output a bounding box around a target object. &
The dataset comprises \textbf{10,567} panoramas within \textbf{90} buildings containing \textbf{4,140} target objects, and \textbf{21,702} crowd-sourced instructions with an average length of \textbf{18} words. & Matterport3D. \\ 
\hline
\textbf{ALFRED} \cite{shridhar2020alfred} & A benchmark for learning a mapping from natural language instructions and egocentric vision to sequences of actions for household tasks. ALFRED includes long, compositional tasks with non-reversible state changes to shrink the gap between research benchmarks and real-world applications. &
ALFRED includes \textbf{25,743} English language directives describing \textbf{8,055} expert demonstrations averaging 50 steps each, resulting in \textbf{428,322} image-action pairs. & AI2-THOR. \\ 
\hline
\textbf{RxR} \cite{ku2020room} & RxR is a larger and multilingual dataset, encompassing English, Hindi, and Telugu, with a greater number of paths and instructions compared to other VLN datasets. It highlights the importance of language in navigation tasks by mitigating existing path biases and encouraging references to observable entities. The dataset was developed to foster advancements in VLN research across different languages. & RxR contains \textbf{126K} instructions covering \textbf{16.5K} sampled guide paths and \textbf{126K} human follower demonstration paths. This dataset is based on building reconstructions from the Matterport3D dataset and viewpoint navigation graphs from the R2R \cite{anderson2018vision} dataset. & Matterport3D.\\ 
\hline
\end{tabularx}
\end{table}

\section{VLN Evaluation Metrics}
\label{section_3}
The main metrics that are used to evaluate the navigation way-finding performance in VLN are as follows: 
\begin{enumerate}
    \item \textbf{Navigation Error (NE)} \cite{anderson2018vision}: is defined as the average shortest-path distance between the agent’s final position and the target destination, effectively measuring how close the agent ends its navigation relative to the goal.
    \item \textbf{Success Rate (SR)}: is the proportion of navigation episodes in which the agent’s final position lies within 3 meters of the target, reflecting its success in reaching the goal. 
    \item \textbf{Success Rate Weighted Path Length (SPL)}: evaluates both accuracy and efficiency by penalizing detours. For each episode, SPL is zero if the agent fails; otherwise, it equals the shortest-path length divided by the agent’s actual path length.
    \begin{equation}
    \text{SPL} = s \cdot \frac{d}{\max(p, d)}
    \label{eq1}
    \end{equation}
    where \textit{s} is 1 if the agent finds any instance of a target, otherwise \textit{s} is 0. \textit{d} is the geodesic distance of the shortest path, and \textit{p} is the distance traveled by the agent. When \textit{s} is 0, SPL will be 0. Otherwise, SPL is in the range of 0 to 1, and a larger SPL means higher efficiency (i.e., shorter path to success).
    \item \textbf{Trajectory Length (TL)}:  denoting the average distance traveled by the agent.
    \item \textbf{Oracle success Rate (OSR)}: the success rate of the agent stopped at the closest point to the goal on its trajectory.
    \item \textbf{Distance to Success (DTS)}: the distance of the agent from the success threshold boundary when the episode ends.
\end{enumerate}

Some other metrics are used to measure the faithfulness of instruction following and the fidelity between the predicted and the ground-truth trajectory, for example: 
\begin{enumerate}
    \item \textbf{Coverage Weighted by Length Score (CLS)} \cite{jain2019stay}: CLS is the product of the Path Coverage (PC) and Length Score (LS) of the agent’s path \textit{\textbf{P}} with respect to reference path \textit{\textbf{R}}: 
    \begin{equation}
        CLS(P, R) = PC(P, R) \cdot LS(P, R)
        \label{eq2}
    \end{equation}
    \item \textbf{Normalized Dynamic Time Warping (nDTW)} \cite{ilharco2019general}: quantifies the alignment between predicted and reference paths by penalizing deviations, thereby providing a robust measure of trajectory fidelity. 
    \item \textbf{Normalized Dynamic Time Warping Weighted by Success Rate (sDTW)} \cite{ilharco2019general}: which penalizes deviations from the ground-truth trajectories and also considers the success rate.
    \item \textbf{Remote Grounding Success rate (RGS)} and \textbf{RGS weighted by Path Length (RGSPL)} are used as whole-task performance metrics to measure the percentage of tasks that correctly locate the target object. 
\end{enumerate}

Evaluation Metric for Instruction generation: BLEU \cite{papineni2002bleu}, CIDEr \cite{vedantam2015cider}, METEOR \cite{banerjee2005meteor}, ROUGE \cite{lin2004rouge}, and SPICE \cite{anderson2016spice} are used. For each navigation path, the metrics are averaged over all the corresponding ground-truth instructions. \textbf{SPICE} is considered the primary metric.


\section{Review of Recent Developments}
\label{section_4}

VLN is a pivotal research domain as it bridges NLU and robotic perception, enabling autonomous systems to interpret and execute high-level human instructions in complex real-world environments. The survey is structured into three sections to provide a systematic view of the advancements in the field. This foundational review offers a structured synthesis of recent progress in VLN, considering some of its challenges, organized around three core thematic pillars that define the field’s current trajectory.

\subsection{Initial Work}
\label{section_4.1}
The concept of VLN  was first formally introduced by Anderson \emph{et al.}\ \cite{anderson2018vision} in 2018. They trained the seq-2-seq model with student-forcing, which achieves promising results in previously explored environments. But the experiments suggested that neural network approaches can strongly overfit to training environments, even with regularization. This made generalizing to unseen environments challenging. They convene a working group to standardize empirical methodologies in 3D mobile navigation research, which has recently seen rapid growth but suffers from disparate task definitions and evaluation protocols. Their report \cite{anderson2018evaluation} consolidates the group’s consensus on problem formulations, generalization objectives, evaluation metrics, and benchmark scenarios to guide and harmonize future navigation studies. 

 Wang \emph{et al.}\ \cite{wang2018look} address the gap between synthetic deep reinforcement learning (DRL) models and real‑world VLN by introducing a hybrid RL framework. The \textbf{look‑ahead module} integrates a policy network with an environment model that predicts next states and rewards, enabling planned exploration. By simulating imagined trajectories, their method enhances scalability and opens avenues for cross‑task transfer in EAI. They also tackle three VLN challenges, cross-modal grounding, sparse feedback, and generalization, via two complementary methods. \textbf{Reinforced Cross-Modal Matching (RCM)} \cite{wang2019reinforced} uses an intrinsic “\textit{matching critic}” reward to align full trajectories with instructions and a reasoning navigator for local grounding. To close the seen–unseen performance gap, \textbf{Self-Supervised Imitation Learning (SIL)} has the agent replay its own successful actions. Both RCM and SIL are modular, model-agnostic, and demonstrate strong generalization in lifelong VLN scenarios.

Chi \emph{et al.}\ \cite{chi2020just} explores various learning strategies to enhance agent interaction at different levels of complexity. They introduce an advanced method using RL with reward shaping, allowing the agent to strategically determine when and where to seek human assistance. Their results demonstrate that the RL-trained agent can effectively adapt to noisy human responses. 
Zhang \emph{et al.}\ \cite{zhang2020diagnosing} investigate various semantic representations that minimize low-level visual details, enabling agents to generalize more effectively to unseen environments. Without altering the baseline model architecture or training process, their semantic features significantly reduce the performance gap between seen and unseen environments across multiple datasets. Majumdar \emph{et al.}\ \cite{majumdar2020improving} introduce \textbf{VLN-BERT}, a vision-language transformer that scores the alignment between navigation instructions and panoramic RGB image sequences. Pretrained on web-scale image-text pairs and fine-tuned on embodied navigation data, this approach highlights the effectiveness of using Internet-scale data to improve VLN performance.

VLN remains challenging due to the complexity of real-world environments and the subtlety of human instructions. \textbf{Object and Action Aware Model (OAAM)} \cite{qi2020object}, which independently models object and action attention to better align visual input and orientation with navigation cues. By decomposing instructions into object and action-specific components and dynamically focusing on segments relevant to the agent's current position, OAAM achieves more precise alignment with candidate viewpoints and enhances navigation accuracy. Whereas the \textbf{Object-aware Vision-and-Language Navigation (OVLN)} \cite{wen2022ovln} model enhances LSTM-based state inference by integrating object features to preserve visual context and improve instruction alignment. Using attention mechanisms, the model captures relational and specific cues from objects, scenes, and directions, forming a visual attention graph for action prediction. Trained on the R2R dataset through a two-stage process: IL and RL followed by data augmentation, OVLN achieves better generalization and effectively addresses the overshoot problem seen in earlier approaches.

Understanding relationships between scenes, objects, and directional cues for interpreting complex navigation instructions is a key factor. To model these interactions, a \textbf{Language and Visual Entity Relationship Graph} \cite{hong2020language} that captures both inter-modal (text-vision) and intra-modal (visual-visual) connections was proposed. A message-passing mechanism propagates information through the graph to inform the agent's actions. While objects serve as visual features in their current approach, future work can leverage them more effectively for tracking progress, localization, and reward shaping through graph-based modeling. Effective planning in VLN demands linking language instructions to an evolving world model and executing long-range exploration with error recovery. Deng \emph{et al.}\ address these challenges with the \textbf{Evolving Graphical Planner (EGP)} \cite{deng2020evolving}, which incrementally builds a dynamic graph from sensory inputs, broadens the agent’s global action space, and performs efficient search over a lightweight proxy representation. Future work can tackle computational efficiency and extend planning horizons for more robust navigation. 

Existing VLN tasks typically rely on navigation graphs, static topological maps of 3D space, leaving open the challenge of how agents construct and update such representations in unfamiliar environments, especially given the difficulties of indoor localization. \textbf{Vision and Language Navigation in Continuous Environments (VLN-CE)} \cite{krantz2020beyond} addresses this by introducing a continuous 3D navigation setting with crowd-sourced instructions, where agents execute low-level actions to follow natural language directions. This setting eliminates several assumptions of prior work, including known environment topology, perfect localization, and short-range oracle guidance. The VLN-CE dataset comprises 4,475 trajectories adapted from the R2R training and validation sets. Two models are introduced: a basic seq-2-seq baseline and an advanced cross-modal attention model. Both agents utilize RGB and depth inputs encoded by pretrained networks designed for image classification and point-goal navigation. VLN-CE serves as the first step toward bridging high-level instruction with low-level control, offering a platform for deeper exploration into integrated, modular learning approaches. They also propose a class of language-conditioned waypoint prediction networks, exploring a range from low-level actions to continuous waypoint predictions \cite{krantz2021waypoint}. Their work emphasizes the need for further research to bridge the gap between topological VLN and continuous VLN-CE, and to strengthen the interface between language understanding and robotic control. The two main shortcomings in current VLN-CE agents are a strict separation between high-level viewpoint planning and low-level motion control, and an over-reliance on RGB/depth data that ignores semantic object attributes vital for assessing navigational feasibility. \textbf{Dual-action module} \cite{zhang2024narrowing} jointly trains agents on both way point selection and physical movement, grounding high-level visual decisions in actual spatial motions. At the same time, it augments way point prediction with rich semantic representations of object properties, enabling the agent to evaluate whether a proposed action is physically possible. 


Generalization to unseen environments remains a major challenge in VLN, with most models showing significant performance drops compared to seen settings. To address this, Tan \emph{et al.}\ \cite{tan2019learning} propose a two-stage training framework. First, the agent is trained using a mix of IL and RL to leverage both off-policy and on-policy benefits. In the second stage, they fine-tune the agent on synthetic unseen (environment, path, instruction) triplets created using a novel \textbf{environmental dropout} technique that simulates unseen environments by selectively masking training data. Whereas Li \emph{et al.}\ address generalizing instruction representations and action decoding, through two effective strategies. They leverage large-scale pretrained language models for better instruction understanding and introduce a stochastic sampling scheme to help agents learn from their own mistakes during sequential decision-making. Their approach, \textbf{PRESS} \cite{li2019robust}, achieves a 6\% absolute improvement in SPL on the R2R benchmark. PRESS’s components are simple yet effective, offering a strong baseline for future VLN models.

Ma \emph{et al.}\ \cite{ma2019self} propose a \textbf{self‑monitoring VLN agent} comprising two modules: (1) a \textbf{visual‑textual co‑grounding} component that identifies the last completed instruction, the next required instruction, and the subsequent movement direction from egocentric images, and (2) a \textbf{progress monitor} that estimates completion of grounded instructions to align navigation actions with overall goal progress. Evaluated on the R2R benchmark, the self‑monitoring paradigm offers a generalizable framework for enhancing decision‑making agents across EAI tasks. They also reframe VLN as graph search, replacing costly beam search with a learned heuristic. They integrate a \textbf{Progress Monitor} \cite{ma2019regretful} trained to estimate goal proximity from grounded language and visual cues, into a greedy best-first search. A \textbf{Regret Module} learns when to backtrack based on progress, while a \textbf{Progress Marker} de-emphasizes visited directions with low estimated progress. This end-to-end agent reduces failed rollbacks and outperforms baselines even when trained on synthetic data. Future work could incorporate efficient exploration strategies and combine goal-driven perception with RL for less structured tasks.  


\textbf{PREVALENT} \cite{hao2020towards} (\textbf{PRE}-TRAINED \textbf{V}ISION-AND-\textbf{L}ANGUAGE BAS\textbf{E}D \textbf{N}AVIGA\textbf{T}OR) is the first pertaining fine‑tuning paradigm for VLN. They self‑supervise on large-scale image‑text‑action triplets to learn generic representations of visual scenes and instructions, which can be integrated into any VLN framework. PREVALENT accelerates learning on new tasks and improves generalization to unseen environments. Evaluations on R2R \cite{anderson2018vision}, CVDN \cite{thomason2020vision}, and HANNA \cite{nguyen2019help} benchmarks confirm that PREVALENT outperforms prior methods.

\subsection{Growth, Diversification, and Mid-Term Advances} 
\label{section_4.2}
Relative directions (e.g., left, right, front, back) and room types (e.g., living room, bedroom) offer crucial semantic and spatial cues for VLN tasks. Addressing this, Qi \emph{et al.}\ \cite{qi2021road} propose the \textbf{O}bject-and-\textbf{R}oom \textbf{I}nformed \textbf{S}equential BER\textbf{T} \textbf{(ORIST)}, which enhances the encoding of instructions and visual information by precisely aligning words with corresponding object regions. Whereas Zhu \emph{et al.}\ \cite{zhu2021soon} propose the \textbf{Scenario Oriented Object Navigation (SOON)} task, requiring agents to locate objects from arbitrary starting points in 3D environments. To support this, they introduce the \textbf{From Anywhere to Object (FAO)} dataset, offering 3K richly annotated natural language instructions. Moving beyond IL, they incorporate RL and develop \textbf{Graph-Based Exploration (GBE)}, which models explored regions as feature graphs for more robust policy learning. Their work presents a promising step toward bridging simulation and real-world navigation challenges.

Zhu \emph{et al.}\ \cite{zhu2021diagnosing} findings reveal that Transformer-based agents exhibit superior cross-modal understanding and stronger numerical reasoning compared to non-Transformer models. However, issues such as imbalanced attention between visual and textual inputs and unreliable cross-modal alignments persist. These insights highlight the need for deeper investigation into the interpretability of neural VLN agents and encourage further research to enhance both task design and agent performance. In that regard, a \textbf{Du}al-scal\textbf{e} Graph \textbf{T}ransformer \textbf{(DUET)} \cite{chen2022think} dynamically builds topological maps for efficient exploration, integrates fine-scale local and coarse-scale global representations via graph transformers, which improves reasoning and language grounding. Despite strong results, challenges remain in generalization to unseen environments and in extending to continuous settings. 

Humans naturally form semantic maps of their environment to navigate using language. \textbf{Semantic Instance Maps (SI Maps)} \cite{nanwani2023instance}, an embedding-free, memory-efficient scene representation for indoor navigation which enables effective language-guided navigation. However, SI Maps currently do not differentiate between instances of the same object, highlighting the potential of integrating 3D instance segmentation for richer semantic mapping. Whereas \textbf{Visual Language Maps} (\textbf{VLMaps}) \cite{huang2023visual} for Robot Navigation, fuses pretrained vision language embeddings with 3D scene reconstructions. VLMaps can be built autonomously from a robot’s video feed using standard exploration, supporting natural language indexing without extra annotations. Future work can refine VLMaps with stronger vision language backbones and adapt them to dynamic scenes with moving objects and people. Effectively anchoring additional modalities like audio for cross-modal reasoning in robotics remains underexplored. Huang \emph{et al.}\ \cite{huang2023audio} address this by introducing \textbf{Audio-Visual-Language Maps} (\textbf{AVLMaps}), a unified 3D representation that stores visual, audio, and language cues. AVLMaps enable robots to locate goals through multimodal queries, with audio inputs improving disambiguation.


Simulation-trained agents rarely leverage mapping strategies vital for real-world navigation. \textbf{Iterative Vision-and-Language Navigation (IVLN)} \cite{krantz2023iterative} is a framework where agents preserve long-term memory across consecutive instruction-following tours. The discrete and continuous \textit{Iterative Room-to-Room (IR2R)} benchmarks, comprising roughly 400 tours across 80 indoor scenes, offer a more realistic platform for advancing embodied navigation research. \textbf{Exploration with Soft Commonsense constraints (ESC)} \cite{zhou2023esc}, which uses a pretrained VLM for prompt-based grounding and a commonsense language model for room–object reasoning, then translates that knowledge into soft logic predicates to guide exploration. ESC achieves significant gains over baselines on the MP3D, HM3D, and RoboTHOR \cite{deitke2020robothor} benchmarks. Future work could enrich this framework with deeper LLM-sourced spatial relations and broaden its application across EAI tasks.

Most existing navigation agents focus solely on translating instructions into actions, offering limited interactivity. Wang \emph{et al.}\ \cite{wang2023lana} address this gap with \textbf{LANA} (a \underline{la}nguage-capable \underline{na}vigation agent), a unified model that both executes human navigation commands and generates route descriptions. LANA employs two encoders (for route and language) whose outputs feed shared decoders for action prediction and instruction generation. This dual-capability framework lays a strong foundation for socially intelligent, trustworthy navigation robots, with further enhancements expected by integrating large-scale pretrained foundation models. Dynamic indoor place recognition faces challenges from lighting shifts and object rearrangements. In that regard,  a \underline{F}oundation-\underline{M}odel \underline{Loc}alization (\textbf{FM-Loc}) \cite{mirjalili2023fm} method that fuses GPT-3 \cite{brown2020language} and CLIP \cite{radford2021learning} to generate semantic descriptors robust to scene geometry and viewpoint changes. FM-Loc automatically selects the most informative landmarks and extends seamlessly to new environments without retraining. Real-world tests in dynamically changing indoor settings validate its strong adaptability and reliability.


\textbf{RobotSlang} \cite{banerjee2021robotslang} is a benchmark of 169 natural language dialogs between a human DRIVER (robot pilot) and a human COMMANDER (navigator). It comprises nearly 5,000 utterances and over 1,000 minutes of robot footage. It defines two tasks, \textit{Localization from Dialog History (LDH)} and \textit{Navigation from Dialog History (NDH)}, and demonstrates that a simulation-trained agent can execute these dialogs on a real robot. Similarly, \textbf{Talk The Walk} \cite{de2018talk} is the first large‑scale dialog dataset grounded in both action and perception. In this task, a “guide” and a “tourist” use natural language to lead the tourist to a target location. \textit{Masked Attention for Spatial Convolutions (MASC)} addresses tourist localization, which grounds utterances in the guide’s map. It significantly improves both emergent and natural language communication and establishes strong baselines on this challenging, open‑ended task.

Traditional VLN agents depend on static environments and expert supervision, hindering real-world deployment. In that regard, \textbf{Human‑Aware VLN} (HA‑VLN) \cite{li2024human} integrates dynamic human activities into navigation tasks, and introduces \textbf{HA3D} simulator, which is built on Matterport3D with moving humans, and \textbf{HA‑R2R}, an extension of R2R annotated with human activity descriptions. To navigate these dynamic scenes, two agents were developed: \textbf{VLN‑CM}, which uses expert‑supervised \textbf{cross‑modal} fusion, and \textbf{VLN‑DT}, a \textbf{decision transformer} trained with non‑expert data. The evaluation using human‑activity aware metrics reveals new challenges in HA‑VLN and highlights the directions for enhancing robustness and sim‑to‑real transfer in populated environments. Whereas \textbf{History Aware Multimodal Transformer (HAMT)}\cite{chen2021history} is the first end-to-end transformer for VLN that replaces recurrent memory with a hierarchical ViT \cite{dosovitskiy2020image} to encode long-horizon history. HAMT processes individual images, captures spatial relations within panoramas, and models temporal links across steps, then fuses these features with instructions and current observations to predict actions. Trained on proxy tasks and fine-tuned with RL, HAMT excels at long-trajectory navigation. Future work may extend HAMT to continuous actions and leverage larger pretraining corpora.

Vision-and-language BERT has boosted many multimodal tasks but struggles in VLN’s partial‑observability setting, which demands history‑dependent reasoning. Considering this, Hong \emph{et al.}\ \cite{hong2021vln} introduce a time‑aware \textbf{recurrent BERT}, augmenting the model with a recurrent module that preserves cross‑modal state across timesteps. Whereas Moudgil \emph{et al.}\ \cite{moudgil2021soat} introduce a transformer‑based VLN agent that leverages two visual encoders, a \textit{scene classifier} for high‑level context and an \textit{object detector} for fine‑grained cues, to align scene descriptions (e.g., “bedroom”) and \textit{object references} (e.g., “green chairs”). By incorporating vision‑language pretraining on large-scale web data, their model achieves a 1.8\% absolute SPL gain on R2R and a 3.7\% absolute SR gain on RxR.

Humans use diverse visual, spatial, and semantic cues when navigating unfamiliar buildings. To endow agents with similar foresight, \textbf{Pathdreamer} \cite{koh2021pathdreamer}, a visual world model that, from past observations, generates high‑resolution 360° RGB, depth, and semantic views at unvisited locations in novel indoor scenes. In uncertain areas, around corners or beyond closed doors, it produces multiple plausible predictions, aiding downstream VLN by providing roughly half the benefit of actual look‑ahead observations. This capability paves the way for model‑based strategies in embodied navigation and object‑search tasks. Whereas Wang \emph{et al.}\ address VLN’s lack of explicit mapping and long-term planning by introducing \textbf{Structured Scene Memory (SSM)} \cite{wang2021structured}, an explicit, compartmentalized memory that disentangles visual and geometric cues into a persistent scene representation. SSM’s \textit{collect–read} controller supports both immediate decisions and iterative, long-range reasoning, while a \textit{frontier-exploration} strategy leverages the full action space of navigable locations for global planning. 


Liang \emph{et al.}\ \cite{liang2022visual} address another challenge of VLN’s large search space and limited generalization by introducing \textbf{ProbES (Prompt-based Environmental Self-exploration)}, a method that eliminates the need for human-labeled navigation data. It leverages CLIP to autonomously sample trajectories and generate structured instructions, creating an in-domain dataset through self-exploration. Instead of conventional fine-tuning, they use prompt-based learning to efficiently adapt language embeddings, enabling rapid cross-domain adaptation. Experiments on R2R and REVERIE show that ProbES improves both performance and generalization, demonstrating the effectiveness of synthesized data and prompt tuning in bridging domain gaps in VLN tasks. Whereas Yu \emph{et al.}\ \cite{yu2023frontier} propose a \textbf{frontier semantic exploration} framework to enhance visual target navigation in large, unknown environments. Traditional methods struggle with complex scene representation and policy learning. Their approach constructs semantic and frontier maps from observations, using DRL to develop a frontier semantic policy that selects frontier cells as long-term goals for efficient exploration. Experiments in Gibson and HM3D environments show improved SR and efficiency over prior map-based methods.


Saha \emph{et al.}\ introduce \textbf{MoViLan} (\underline{Mo}dular \underline{Vi}sion \underline{Lan}guage Navigation) \cite{saha2022modular}, a modular framework for executing visually grounded household instructions. Unlike end-to-end VLMs that falter on long‑horizon, compositional tasks with diverse objects and irreversible changes, MoViLan trains VLMs separately, eliminating the need for aligned trajectory data. It combines a geometry‑aware mapper for cluttered indoor scenes with a generalized language model for household directives, achieving markedly higher success rates on long‑horizon tasks in the ALFRED benchmark. Li \emph{et al.}\ introduce \textbf{PANOGEN} \cite{li2023panogen}, a text-conditioned panorama generator that uses recursive diffusion outpainting on MP3D captions to create diverse, semantically coherent 360° environments. By (1) synthesizing instruction–path pairs for pre-training and (2) augmenting visual inputs during fine-tuning, PANOGEN boosts VLN generalization settings on R2R, R4R, and CVDN, with especially strong gains on under-specified CVDN instructions.

Dorbala \emph{et al.}\ \cite{dorbala2022clip} explore the potential of VLMs, particularly CLIP, for zero-shot VLN using natural language referring expressions, moving beyond prior work based on simple class-based instructions. Without dataset-specific finetuning, their CLIP-based agents: \textbf{CLIP-Nav} and \textbf{Seq CLIP-Nav} demonstrate strong generalization and consistent performance across environments on the REVERIE dataset, outperforming supervised baselines in SR, SPL, and Relative Change in Success (RCS). Their results highlight CLIP’s ability to make accurate sequential navigation decisions in zero-shot settings. Future directions can include evaluating cross-dataset performance, incorporating dialog, improving backtracking with meta-learning, and studying human navigation patterns in Virtual Reality (VR) to inform model behavior. Whereas Wang \emph{et al.}\ introduce \textbf{VXN} \cite{wang2022towards}, a large-scale 3D dataset unifying four navigation tasks (image, object, audio-goal, and VLN) in continuous, multimodal environments, and \textbf{VIENNA}, a single transformer-based agent that tackles all four tasks via a unified parse-and-query framework. VIENNA encodes each target with task-specific embeddings into dynamic goal vectors, which it refines during navigation and uses to attend over episodic memory for decision making. Experiments show that, compared to training separate agents, VIENNA matches or exceeds single-task performance while greatly simplifying deployment.

Li \emph{et al.}\ introduce \textbf{ENVEDIT} \cite{li2022envedit}, a data-augmentation technique that edits existing VLN environments along three axes: style, object appearance, and object classes. Training on these varied settings reduces overfitting and boosts generalization. On R2R and RxR benchmarks, ENVEDIT improves all metrics for both pretrained and from-scratch agents. Moreover, ensembling agents trained on differently edited environments yields further gains, demonstrating the complementarity of the edit methods. Future work may leverage more advanced image-translation models and extend ENVEDIT to other embodied tasks. Whereas \textbf{ScaleVLN} \cite{wang2023scaling} is a large-scale data augmentation paradigm for VLN that leverages over 1,200 photorealistic HM3D and Gibson scenes to synthesize 4.9 million instruction–trajectory pairs from web resources. They systematically evaluate the impact of graph quality, image fidelity, and pretraining strategies on agent performance. Simple imitation learning on this dataset boosts an existing agent’s R2R single-run success rate and shrinks the seen–unseen generalization gap. ScaleVLN delivers results on CVDN, REVERIE, and R2R-CE, offering a practical blueprint for large-scale VLN data generation and utilization.

Gao \emph{et al.}\ tackle REVERIE, which requires goal-driven exploration and remote object localization from high-level instructions. They introduce \textbf{Cross-modality Knowledge Reasoning (CKR)} \cite{gao2021room}, a transformer-based model that builds scene memory tokens for informed exploration. CKR’s \textit{Room-and-Object Aware Attention (ROAA)} extracts room and object cues from text and vision, while the \textit{Knowledge-enabled Entity Relationship Reasoning (KERR)} module uses commonsense graphs to reason over room–object correlations for action selection. Whereas Lin \emph{et al.}\ introduce a \textbf{scene-intuitive} agent \cite{lin2021scene} for REVERIE using a two-stage training pipeline. First, the model learns cross-modal alignment via \textbf{Scene Grounding} (where to stop) and \textbf{Object Grounding} (what to attend) tasks. Second, a memory-augmented attentive decoder fuses these grounded representations with past experiences to generate actions. Without additional bells and whistles, this approach performs well, highlighting the value of explicit grounding and memory in high-level instruction following.

Qiao \emph{et al.}\ introduce \textbf{HOP} \cite{qiao2022hop}, a history-and-order aware pretraining paradigm for VLN that enhances spatio-temporal grounding and action foresight. Building on Masked Language Modeling (MLM) and Trajectory-Instruction Matching (TIM), they add \textit{Trajectory Order Modeling} and \textit{Group Order Modeling} to capture temporal structure, plus \textit{Action Prediction} with History to condition decisions on past observations. While computationally intensive and currently focused on indoor, graph-based environments, HOP paves the way for more efficient architectures and extensions to outdoor or continuous settings. They extend and propose \textbf{HOP+} \cite{qiao2023hop+}, a VLN pretraining–fine-tuning framework that integrates historical context and temporal reasoning. In addition to MLM and TIM, HOP+ introduces three VLN-specific pretraining tasks: \textit{Action Prediction with History (APH), Trajectory Order Modeling (TOM), and Group Order Modeling (GOM)}. To align pretraining with fine-tuning, they employ an external memory network that selectively retrieves historical features for action decisions without significant overhead. HOP+ sets new results on R2R, REVERIE, RxR, and NDH, validating its effectiveness in enhancing temporal grounding and decision-making.

Majumdar \emph{et al.}\ \cite{majumdar2022zson} propose a scalable, zero-shot method for open-world \textit{object-goal navigation} (ObjectNav) by training agents on an \textit{image-goal navigation} (ImageNav) task using a multimodal semantic embedding space. This allows agents to interpret free-form language goals (e.g., “find a sink”) without requiring demonstrations or ObjectNav-specific rewards. Their \textbf{SemanticNav} agents generalize well across diverse environments (e.g., HM3D, MP3D, Gibson) and outperform prior zero-shot methods by 4.2–20\% in SR. The agents can also handle compound instructions involving room context. Key success factors include visual encoder pretraining and diverse training environments. However, agents may fail when objects appear in typical locations due to training data biases. Future work could use language prompts to guide exploration in such cases. Whereas Chen \emph{et al.}\ \cite{chen2023object} propose an implicit spatial mapping approach for object-goal navigation that addresses the limitations of classical mapping and end-to-end methods. Their model uses a transformer to recursively update the map with new observations and incorporates auxiliary tasks, explicit map reconstruction, visual feature prediction, and semantic labeling to enhance spatial reasoning. The method achieves good performance on the MP3D dataset, generalizes to HM3D, and demonstrates effective real-world deployment with minimal fine-tuning.

Gao \emph{et al.}\ present \textbf{Adaptive Zone-aware Hierarchical Planner (AZHP)} \cite{gao2023adaptive}, a novel hierarchical policy framework for VLN. Unlike traditional single-step planning methods, AZHP decomposes navigation into two asynchronous levels: high-level subgoal planning and low-level execution. A \textit{State-Switcher Module (SSM)} coordinates these phases. At the high level, \textit{Scene-aware Zone Partition (SZP)} dynamically segments the environment into zones, and \textit{Goal-oriented Zone Selection (GZS)} identifies the target zone for each subgoal. At the low level, the agent executes multi-step navigation within the selected zone. The framework is trained using \textit{Hierarchical Reinforcement Learning (HRL)} augmented with auxiliary objectives and \textit{curriculum learning}. AZHP experimented on REVERIE, SOON, and R2R benchmarks, demonstrating the effectiveness of hierarchical planning in VLN. The other works like \textbf{Robo-VLN} \cite{irshad2021hierarchical} is a more realistic VLN framework set in continuous 3D reconstructed environments with longer trajectories, continuous control, and obstacle challenges. The \textit{Hierarchical Cross-Modal (HCM)} agent is proposed, which leverages layered decision-making, modular training, and the decoupling of reasoning from imitation.


Traditional VLN benchmarks focus exclusively on ground-based agents, overlooking the unique challenges of aerial navigation, namely altitude control and 3D spatial reasoning. To bridge this gap, Liu \emph{et al.}\ \cite{liu2023aerialvln} introduce \textbf{AerialVLN}, a city-scale, UAV-based VLN benchmark. It supports long-distance 3D path planning in novel outdoor environments and invites future research into extended-horizon action learning under sparse rewards. In the same direction, a zero‑shot aerial VLN framework that uses an LLM for action prediction, powered by a \textbf{Semantic‑Topo‑Metric Representation (STMR)} \cite{gao2024aerial}, extracts instruction‑relevant semantic masks of landmarks, projects them onto a top‑down map of landmark locations, and encodes this as a distance‑metric matrix prompt. The LLM then leverages this spatial representation to predict navigation actions.

Initially, Anderson \emph{et al.}\ \cite{anderson2021sim} evaluate \textbf{sim‑to‑real transfer} of a VLN agent by deploying a simulation-trained model on a physical robot. They introduce a subgoal model to convert high‑level discrete actions into nearby continuous waypoints and apply domain randomization to bridge visual gaps. To compare performance, they annotate a 325m² office with 1.3 km of instructions and its simulated replica. Results show robust transfer when an occupancy map and navigation graph are pre‑annotated, but performance drops sharply without prior mapping. Whereas Wang \emph{et al.}\ \cite{wang2024sim} introduce a sim‑to‑real transfer method that endows monocular robots with panoramic traversability perception and semantic understanding. They generate a semantic traversable map to predict agent‑centric waypoints and use 3D feature fields to synthesize novel views at those points. This expands a robot’s field of view and markedly boosts navigation performance. Their system surpasses prior monocular VLN methods on R2R‑CE and RxR‑CE benchmarks in simulation and demonstrates strong results in real-world tests.

\subsection{Current Breakthroughs and Latest Contributions}
\label{section_4.3}


Zero-Shot Object Navigation (ZSON) enables agents to locate open-vocabulary objects in unfamiliar environments without prior training. \textbf{Zero-shot Interactive Personalized Object Navigation} (ZIPON) \cite{dai2024think}, where robots navigate to personalized goal objects through dialogue with users. \textbf{ORION} (\textbf{O}pen-wo\textbf{R}ld \textbf{I}nteractive pers\textbf{O}nalized \textbf{N}avigation), leverages LLMs to coordinate perception, navigation, and communication modules through sequential decision-making. While balancing task success, efficiency, and interaction remains a challenge, this work marks a significant step toward adaptive, conversational agents for personalized human-robot collaboration. Similarly, \textbf{ApexNav} \cite{zhang2025apexnav} is also a ZSON framework designed for efficient and reliable navigation in unfamiliar environments. To balance exploration strategies, this adaptively switches between semantic reasoning and geometry-based navigation based on the strength of semantic cues. For improved reliability, it incorporates a target-centric semantic fusion mechanism that maintains long-term memory of the target and visually similar objects, reducing false detections. It outperforms existing methods on HM3Dv1, HM3Dv2, and MP3D datasets in SR and SPL metrics, with ablation and real-world experiments confirming its effectiveness and practical applicability.

Hou \emph{et al.}\ present \textbf{ELA-ZSON} \cite{hou2025ela} (Efficient Layout-Aware ZSON Agent with Hierarchical Planning), a layout-aware ZSON method for complex indoor environments. It combines hierarchical planning using a global topological map with local scene memory, guided by an LLM agent, enabling efficient navigation without human input, reward engineering, or extensive training. It achieves strong results on the MP3D benchmark with 85\% SR and 79\% SPL. Its effectiveness is further validated through simulations and real-world deployment. Limitations include suboptimal use of local scene memory and a lack of dynamic scene updates. Similarly, \textbf{hierarchical, semantic knowledge-based object search} framework \cite{zhang2023hierarchical} that enables robots to emulate reasoning and incorporates prior knowledge linking rooms to typical objects, enhancing the robustness of the semantic framework. A set of rules is then introduced for deploying and updating this knowledge, enabling a heuristic search strategy that guides robots to target objects more quickly. Future directions can include developing real-time adaptive search strategies for more dynamic and responsive object search. Whereas \textbf{HOV-SG}  \cite{werby2024hierarchical}, a \textbf{H}ierarchical \textbf{O}pen-\textbf{V}ocabulary 3D \textbf{S}cene \textbf{G}raph framework for language-guided robot navigation, leverages vision foundation models, generates open-vocabulary 3D segment maps and builds a multi-level scene graph encompassing floors, rooms, and objects, each enriched with open-vocabulary features. A language-grounded navigation module, powered by GPT-3.5 \cite{kuan2024understanding}, decomposes complex queries (e.g., \textit{“find the toilet in the bathroom on floor 2”}) into structured sub-queries across the hierarchy. This approach demonstrates promising generalization and scalability, with future directions aimed at dynamic scene representation and the integration of reactive embodied agents for enhanced reasoning and interaction. 

Zhang \emph{et al.}\ propose \textbf{TriHelper} \cite{zhang2024trihelper}, a ZSON framework that addresses key challenges like collision, inefficient exploration, and target misidentification through three specialized modules: \textit{Collision Helper, Exploration Helper,} and \textit{Detection Helper}. Unlike prior holistic approaches, TriHelper provides dynamic, targeted assistance throughout navigation. Experiments on HM3D and Gibson datasets show that TriHelper outperforms existing baselines in SR and exploration efficiency. Ablation studies confirm the contribution of each module, emphasizing the value of modular, challenge-specific support in advancing Zero-Shot ObjectNav and embodied AI. Whereas Cai \emph{et al.}\ propose \textbf{CL-CoTNav} \cite{cai2025cl}, a VLM-driven ObjectNav framework that enhances generalization in unseen environments through structured reasoning and closed-loop feedback. Unlike traditional end-to-end methods, this leverages Hierarchical Chain-of-Thought (H-CoT) prompting, fine-tuned on multi-turn QA data from human trajectories to simulate human-like iterative reasoning. A Closed-Loop H-CoT mechanism further improves robustness by weighting training data based on detection and reasoning confidence. Experiments in AI Habitat show CL-CoTNav significantly outperforms several baselines. While effective, the model depends on IL, prompting future directions in offline and online RL to enhance adaptability and scalability in real-world settings.

Object tracking and following are vital capabilities for a wide range of robotic applications, including automation, logistics, healthcare, and security. \textbf{Follow Anything (FAn)} \cite{maalouf2024follow} is a real-time robotic system capable of detecting, tracking, and following arbitrary objects using multimodal, open-vocabulary queries (text, images, or clicks), was built on powerful foundation models like CLIP \cite{radford2021learning} and DINO \cite{caron2021emerging}. FAn operates beyond training-time constraints, enabling generalization to unseen object classes at inference. Demonstrated on an aerial vehicle, FAn showcases robust, real-time object following in dynamic environments, highlighting the growing potential of foundation models in practical, multi-modal robotic systems. Whereas Hong \emph{et al.}\ introduce \textbf{VLN with Multi-modal Prompts (VLN-MP)} \cite{hong2024only}, which enriches traditional VLN by combining text instructions with optional image prompts, such as exact or similar landmark pictures, to resolve visual ambiguities. Their benchmark includes: a training-free pipeline that converts prose into multi-modal directives, four downstream datasets, and an \textit{Multi-modal Prompts Fusion (MPF)} module for seamless integration with existing VLN architectures. Across R2R, RxR, REVERIE, and CVDN, visual prompts boost navigation accuracy while retaining full compatibility with text-only inputs, demonstrating VLN-MP’s versatility and practical value.

People with visual impairments often face challenges in spatial understanding and navigation. \textbf{DRAGON} \cite{liu2024dragon} is a dialogue-enabled robot guide that integrates semantic understanding with physical guidance and allows users to interact through natural language. Built using a modular pipeline, it performs effectively in real-world settings, though it currently relies on high-end sensors and rule-based dialogue systems. To enhance scalability and adaptability, future work can aim to replace fixed rules with learning-based policies (e.g., LLMs), enrich environmental understanding through object relationships and multi-modal sensing. DRAGON highlights the promise of VLMs in assistive robotics, paving the way for more interactive and intelligent navigation aids. Whereas Cheng \emph{et al.}\ introduce \textbf{NaVILA} \cite{cheng2024navila}, a two-level framework for legged-robot VLN that bridges high-level language reasoning and low-level locomotion. Rather than mapping instructions directly to joint controls, NaVILA’s Vision-Language Action (VLA) module outputs mid-level, language-based actions (e.g., “\textit{move forward 75 cm}”), which a visual RL locomotion policy then executes. This design enhances robustness and generalization, outperforming prior methods by 17\% on standard VLN benchmarks, including real-robot tests in cluttered scenes.

\textbf{Safe-VLN} \cite{yue2024safe} is a collision-aware navigation framework featuring two key components: a \textbf{waypoint predictor}, which uses simulated 2D LiDAR occupancy masks to avoid obstacle-prone areas, and a \textbf{navigator}, which implements a \textit{‘reselection after collision’} strategy to prevent repeated collisions. Through a detailed classification of collision scenarios, Safe-VLN improves robustness in navigation and significantly reduces collision rates. Future work can include deploying Safe-VLN on physical robots and enhancing sim-to-real transfer through improved perception and real-world data augmentation. Similarly, \textbf{Human-Aware VLN (HA-VLN)} \cite{dong2025ha} is a unified benchmark for VLN that integrates both discrete and continuous paradigms with explicit social-awareness constraints. Key contributions include a standardized task definition incorporating personal-space considerations, an upgraded \textit{HAPS 2.0} dataset with realistic multi-human dynamics and motion–language alignment, and enhanced simulators for diverse indoor and outdoor settings. Evaluated on over 16,000 human-centric instructions, HA-VLN reveals that multi-human interactions and partial observability significantly challenge existing VLN agents. Real-world experiments validate sim-to-real transfer, and a public leaderboard supports standardized evaluation. This work promotes socially aware, safe, and effective navigation in human-populated environments.

Ensuring VLN agents faithfully follow instructions typically requires complex history‑encoding modules. \textbf{VLN‑GPT} \cite{hanlin2024vision} uses a GPT‑2 decoder to capture trajectory dependencies directly from the action sequence, eliminating separate encoders. They split training into offline imitation pretraining and online RL fine‑tuning for targeted optimization. On standard VLN benchmarks, it outperforms more complex encoder‑based models. Whereas \textbf{CONSOLE} (\textbf{CO}rrectable La\textbf{N}dmark Di\textbf{S}c\textbf{O}very via \textbf{L}arge Mod\textbf{E}ls)\cite{lin2024correctable} is a VLN framework that reframes navigation as sequential open-world landmark discovery. It leverages ChatGPT to supply commonsense landmark co-occurrences and employs CLIP to detect these landmarks in the environment. A learnable co-occurrence scorer then refines ChatGPT’s priors using actual observations. Enhanced landmark features feed into any VLN agent’s decision process. Across R2R, REVERIE, R4R, and RxR benchmarks, especially in unseen environments, CONSOLE consistently outperforms strong baselines. This work showcases how large models’ world knowledge can be harnessed to boost embodied navigation.



Chen \emph{et al.}\ \cite{chen2021topological} propose a \textbf{modular VLN} framework inspired by robotics, using topological maps to overcome the limitations of end-to-end models in freely traversable environments. Their method employs attention-based planning over a topological map derived from natural language instructions, followed by low-level action execution via a robust controller. This approach enables interpretable planning and demonstrates intelligent behaviors like backtracking. Their work suggests promising directions for real-world deployment and test-time map construction, aiming to advance robust, communicative robot systems. Most VLN pretraining uses discrete panoramas, forcing models to infer spatial relations from fragmented, redundant views. Whereas An \emph{et al.}\ introduce \textbf{ETPNav} \cite{an2024etpnav}, a framework that decouples long‑range planning from low‑level control. ETPNav builds an online topological map by self‑organizing predicted waypoints along the agent’s path, then uses a transformer‑based cross‑modal planner to generate high‑level routes from this map and language instructions. A trial‑and‑error heuristic controller ensures obstacle avoidance during execution.

\begin{figure}[htbp]
    \centering
    \Description{BEVBert Architecture.}
    \includegraphics[width=\linewidth]{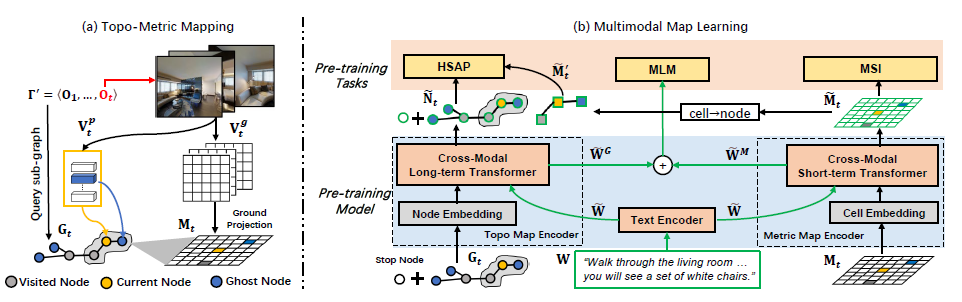}
    \captionsetup{justification=justified}
    \caption{\small BEVBert Architecture\cite{an2022bevbert} .}
    \label{fig:BEVBert Architecture}
\end{figure}

To improve spatial reasoning, \textbf{BEVBert}\cite{an2022bevbert}, a \textbf{map-based pretraining approach}, constructs a local metric map to merge incomplete observations and eliminate duplicates, alongside a global topological map to capture long‑range dependencies. A multimodal pretraining framework then learns spatially aware map representations, enhancing cross‑modal reasoning for language‑guided navigation. It is a hybrid map-based pre-training framework comprising two modules (figure ~\ref{fig:BEVBert Architecture}): \textit{topo-metric mapping} and \textit{multimodal map learning}. The mapping module generates an offline hybrid map from sampled expert trajectories, while the learning module performs map–instruction interactions to pre-train multimodal map representations. The pre-trained model is then fine-tuned for sequential action prediction using online-constructed maps.

\textbf{Memory-Maze} \cite{kuribayashi2024memory} is a virtual maze environment paired with route instructions crowd-sourced both on-site (longer, more varied, and error-filled) and online. An LLM-driven VLN agent that translates these real-world instructions into Python control code was proposed to outperform several baselines without additional training or prebuilt maps. Their results highlight a gap between conventional VLN benchmarks and practical deployment, underscoring the need for adaptive map representations and interactive error-correction modules in future systems. Whereas \textbf{StratXplore} \cite{gopinathan2024stratxplore} is a memory-based, mistake-aware strategy for VLN that enhances error recovery by selecting optimal “frontier” viewpoints, recently seen but unvisited locations that align with instructions. Unlike back-tracking methods, it uses stored action and viewpoint features to monitor progress, ensure task conformity, seek novel views, and prioritize corrective steps. Evaluated on R2R and R4R, it significantly improves SR in unseen environments. Future work can integrate mistake detection directly into the planner for more efficient recovery.

\textbf{Vision-Language Frontier Maps (VLFM)} \cite{yokoyama2024vlfm} is a zero-shot navigation framework inspired by human reasoning, designed to guide robots toward unseen semantic objects in unfamiliar environments. It constructs occupancy maps from depth data to identify exploration frontiers and uses VLMs for semantic guidance. While current assumptions include visible target objects at default camera height, future improvements may involve active exploration strategies, enhanced prompt engineering, refined value maps, and more robust semantic tracking, paving the way for long-horizon and multitask navigation. Similarly, \textbf{PixNav} \cite{cai2024bridging} introduces a pure RGB–based zero‑shot navigation policy that bridges foundation models and embodied agents by using pixel‑level goals. For long‑horizon navigation, an LLM‑based planner leverages commonsense object–room relations to select waypoints. Evaluations in photorealistic simulators and real environments confirm PixNav’s robustness and generalization. Future work can explore larger, diverse navigation datasets and integrate vision–language planners for improved long‑term performance. 

Yue \emph{et al.}\ propose the \textbf{Multi-level Fusion and Reasoning Architecture (MFRA)} \cite{yue2025think} to improve VLN by enabling agents to better integrate visual input, language instructions, and navigation history. MFRA employs a hierarchical fusion mechanism to combine features from low-level visual cues to high-level semantics across modalities, and a reasoning module that uses instruction-guided attention and dynamic context to infer actions. Evaluated on REVERIE, R2R, and SOON benchmarks, it outperforms well, highlighting the effectiveness of multi-level fusion in enhancing VLN performance. Some work highlights critical security concerns in VLM-powered VLN systems and provides a foundation for improving robustness in real-world deployments. In that direction, \textbf{Adversarial Object Fusion (AdvOF)} \cite{xie2025disrupting} is a novel attack framework that targets VLN agents in service-oriented settings by generating adversarial 3D objects. AdvOF aligns 2D and 3D object positions, then optimizes adversarial objects through collaborative learning and multi-view fusion with weighted importance. This approach effectively disrupts agent performance under adversarial conditions while minimally affecting normal navigation.

\textbf{COSMO} (\textbf{CO}mbination of \textbf{S}elective \textbf{M}em\textbf{O}rization) \cite{ramrakhya2025grounding} is a lightweight yet high-performing architecture for VLN, addressing the rising computational costs of transformer-based models augmented with external knowledge or maps. COSMO integrates transformer and state-space modules, incorporating two VLN-specific components: \textit{Round Selective Scan (RSS)} for efficient intra-modal interactions within scans, and \textit{Crossmodal Selective State Space (CS3)} for enhanced cross-modal reasoning via a dual-stream architecture. Evaluations on REVERIE, R2R, and R2R-CE benchmarks show that COSMO achieves competitive performance, especially on long instructions, while significantly reducing computational overhead. Whereas Cui \emph{et al.}\ propose \textbf{Fine-grained Cross-modal Alignment (FCA-NIG)} \cite{cui2025generating}, a generative framework addressing the lack of fine-grained cross-modal annotations in VLN. Existing datasets emphasize global instruction-trajectory alignment, overlooking sub-instruction and entity-level cues crucial for precise navigation. FCA-NIG constructs dual-level annotations, sub-instruction-to-sub-trajectory and entity-to-landmark, by segmenting trajectories and using GLIP \cite{li2021grounded}, OFA \cite{wang2022ofa}, and CLIP \cite{radford2021learning} to generate and align instructions. This process yields the \textit{FCA-R2R} dataset, the first large-scale resource with fine-grained alignments. Training SoTA agents (e.g., SF \cite{fried2018speaker}, EnvDrop \cite{tan2019learning}, RecBERT \cite{hong2021vln}, HAMT \cite{chen2021history}) on FCA-R2R significantly improves navigation accuracy and generalization. The framework enhances decision-making and interpretability, advancing scalable VLN training without manual annotation.

\textbf{NaVid} \cite{zhang2024navid} is a video-based VLM for VLN that navigates using only monocular RGB streams, no maps, odometry, or depth sensors. By encoding spatio-temporal context with special tokens, NaVid learns from 510k in-domain navigation samples and 763k web-sourced image–text pairs. It achieves robust Sim-to-Real transfer in both simulated and real environments. While NaVid’s high computational cost and limited long-horizon context pose challenges, future work can explore action-chunking, larger backbones, and extensions to mobile manipulation tasks. Whereas \textbf{Re}trieval-augmented \textbf{M}emory for \textbf{Em}bodied \textbf{R}obots \textbf{(ReMEmbR)} \cite{anwar2024remembr} addresses long‑horizon video question answering for embodied robots by combining a VLM-built memory store with an LLM‑driven query phase. Evaluated on \textbf{NaVQA}, a new dataset of spatial, temporal, and descriptive questions over extended navigation videos, ReMEmbR outperforms LLM/VLM baselines with low latency. This retrieval‑augmented approach enables robots to leverage growing deployment histories for effective, real‑time reasoning in complex environments.

\section{VLN based Multi-robot systems and Human-Robot Interaction}
\label{section_5}

As we know, visual navigation is a fundamental capability for household service robots, and growing task complexity necessitates effective communication and coordination among multiple agents. While LLMs have demonstrated strong reasoning and planning abilities in embodied contexts, their use for collaborative multi-robot navigation in household environments remains underexplored. Wu \emph{et al.} introduces \textbf{CAMON}\cite{wu2024camon}(Cooperative Agents for Multi-Object Navigation), a fully decentralized framework for cooperative multi-object navigation that leverages LLM-enabled communication. By employing a communication-triggered dynamic leadership structure, CAMON enables efficient task allocation and rapid team consensus with minimal communication overhead. The proposed scheme improves navigation performance and ensures conflict-free collaboration, even as team size increases. Despite its effectiveness, the current approach faces limitations in handling dynamic objects, such as humans or pets, and is restricted to single-floor navigation. These constraints can be addressed through extended perception strategies and cross-floor coordination modules. Future research can focus on integrating communication, navigation, and manipulation to support more complex multi-robot tasks. Similarly, Liu \emph{et al.}\ \cite{liu2022multi} introduce multi‑agent visual semantic navigation, where agents coordinate under limited communication to find multiple targets. Their hierarchical decision framework leverages semantic maps, scene priors, and inter‑agent messaging to guide exploration. Experiments in unseen environments, both with familiar and novel objects, demonstrate superior accuracy and efficiency compared to single-agent baselines. As future work, we can further study the collaboration and communication in relevant embodied tasks such as multi-agent task assignment.

\begin{figure}[htbp]
    \centering
    \Description{Artist impression of VLN in MRS.}
    \begin{subfigure}[b]{0.58\textwidth}
        \centering
        \includegraphics[width=\textwidth]{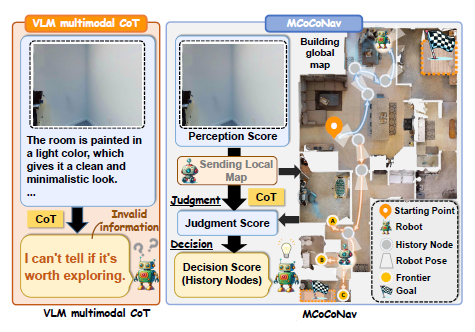}
        \caption{Illustrations show multimodal CoT reasoning in VLMs and cross-image multimodal CoT reasoning in MCoCoNav. By leveraging cross-image reasoning, MCoCoNav enables robots to jointly interpret multiple scene perspectives and the global semantic map, supporting effective zero-shot multi-robot semantic navigation.}
        \label{fig:sub1}
    \end{subfigure}
    \hfill
    \begin{subfigure}[b]{0.38\textwidth}
        \centering
        \includegraphics[width=\textwidth]{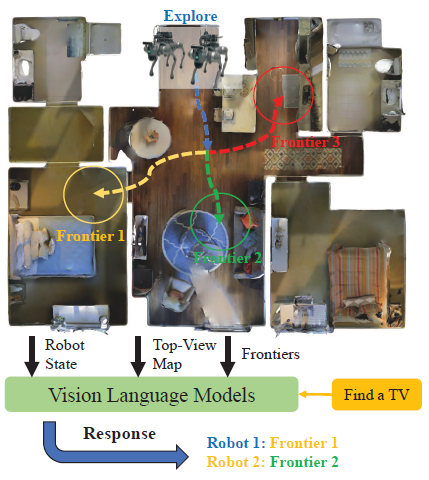}
        \caption{Example of two-robot visual target navigation: When several unexplored frontiers are identified, the vision-language model allocates frontier goals to each robot according to their current observations and the target object.}
        \label{fig:sub2}
    \end{subfigure}
    \caption{(a) MCoCoNav scenario \cite{shen2025enhancing} and (b) Co-NavGPT scenario \cite{yu2023co}}
    \label{fig4}
\end{figure}

\begin{figure}[htbp]
    \centering
    \Description{Artist impression of VLN in MRS.}
    \includegraphics[width=\linewidth]{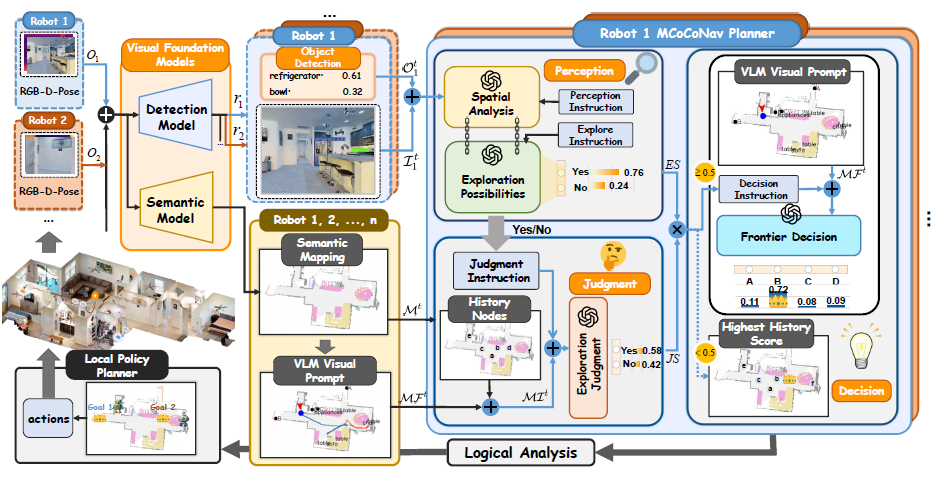}
    \captionsetup{justification=justified}
    \caption{\small Components of MCoCoNav\cite{shen2025enhancing}: The architecture comprises Visual Foundation Models, an MCoCoNav Planner, and a Local Policy Planner. The core MCoCoNav Planner integrates three modules: Perception, Judgment, and Decision.}
    \label{fig5}
\end{figure}

Chen \emph{et al.}\ \cite{chen2024scalable} evaluated \textbf{centralized, decentralized, and hybrid communication} schemes across several coordination tasks, finding that a hybrid approach consistently maximizes task success and gracefully scales to larger teams. Looking ahead, developing hierarchical structures of specialized robot subgroups and integrating emerging multi-modal (particularly visual) models could revolutionize communication paradigms and optimization strategies for large-scale multi-agent systems. Cooperative semantic navigation is vital for home-service MRS, yet centralized planners hinder efficiency, and decentralized schemes often ignore communication overhead (figure~\ref{fig:sub1}). In that regard, Shen \emph{et al.}\ introduce \textbf{MCoCoNav} \cite{shen2025enhancing} (Multimodal Chain-of-Thought Co-Navigation), a modular framework (figure~\ref{fig5}) that uses multimodal chain-of-thought (CoT) \cite{wei2022chain} planning, combining visual inputs with VLM-derived probabilistic scores and a shared semantic map to coordinate exploration while limiting communication costs. Evaluations on HM3D\_v0.2 \cite{ramakrishnan2021habitat} and MP3D \cite{chang2017matterport3d} confirm its enhanced efficiency and robust performance.

An \emph{et al.}\ \cite{an2024scalable} present a decentralized multi-agent RL framework for scalable \textbf{multi-robot, multi-goal (MRMG)} navigation. Their permutation invariant policy trained model-free in simulation, handles varying numbers of robots and targets zero-shot, avoiding order bias and fixed capacities. Compared to a non-invariant baseline, it boosts success by 10.3\% and solves near-optimal MRMG tasks two orders of magnitude faster than centralized optimization. Deployed on wheeled-legged quadrupeds in both simulation and real environments, the approach dynamically prioritizes peers and goals, generalizing to unseen team sizes. Future work can extend to heterogeneous platforms and integrate onboard semantic vision. Similarly, \textbf{InsightSee} \cite{zhang2024insightsee} is a multi-agent framework designed to improve VLM interpretability in complex visual scenarios. It integrates a description agent, two reasoning agents, and a decision agent to enhance visual information processing.

Roman \emph{et al.}\ introduce a two‑agent framework: one navigator and one guider, to advance language‑guided robots beyond passive instruction following. Drawing on Theory of Mind, their \textbf{Recursive Mental Model (RMM)}  \cite{roman2020rmm} has each agent simulate the other: the navigator predicts guider responses to candidate questions, while the guider anticipates the navigator’s actions to craft answers. Progress toward goals serves as an RL reward, shaping navigation, question, and answer generation. RMM enhances generalization in novel environments and offers a blueprint for interactive, task‑oriented agent communication in human–robot collaboration. Whereas \textbf{CoNav} \cite{li2024conav} is a benchmark for collaborative navigation where robots anticipate human goals by observing realistic, diverse human movements in 3D environments. They generate these human trajectories via an LLM-driven animation framework conditioned on textual descriptions and environmental context, compatible with existing simulators. They develop an intention-aware agent that predicts both long and short-term human goals from panoramic observations to guide its own path. CoNav’s results showing improved collision avoidance and proactive following highlight the need for intent-based models and lay the groundwork for advanced human–robot teamwork.

\begin{figure}[htbp]
    \centering
    \Description{Co-NavGPT.}
    \includegraphics[width=\linewidth]{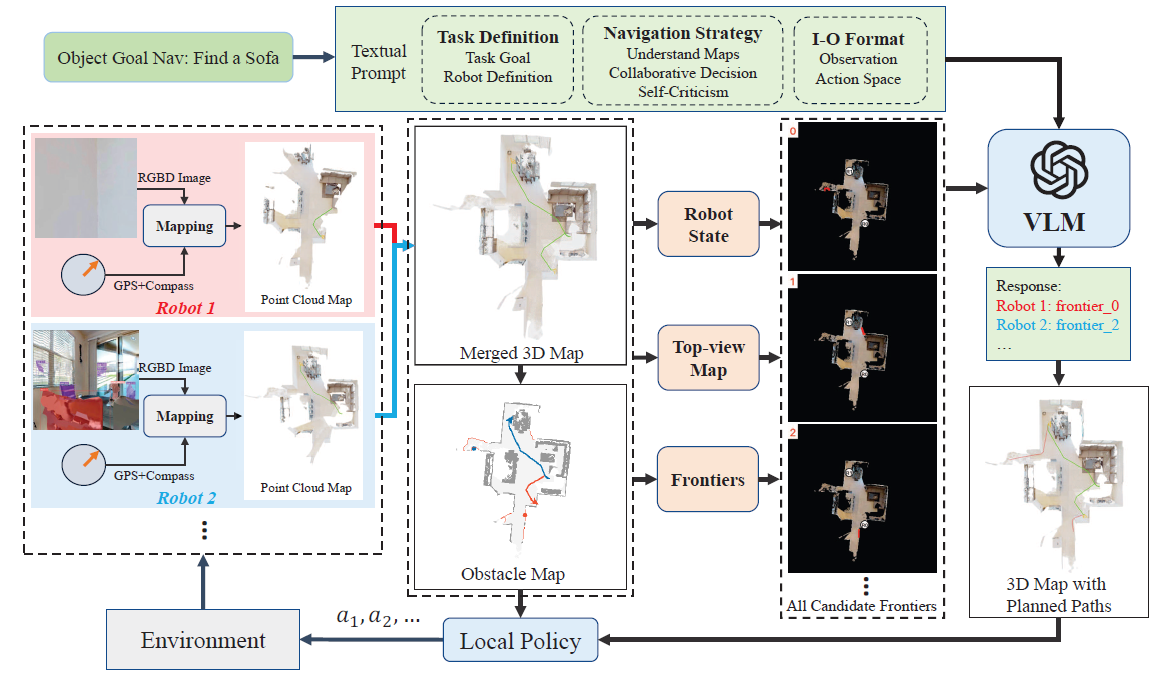}
    \captionsetup{justification=justified}
    \caption{\small Co-NavGPT\cite{yu2023co}: multi-robot navigation framework integrates perception, mapping, and language-based planning for coordinated exploration. Each robot converts RGB-D inputs into local point clouds, which are fused into a global 3D map. This map, along with robot states and frontier candidates, is encoded into structured prompts for a vision-language model that functions as a global planner, allocating exploration goals. Local navigation policies then generate collision-free paths toward these goals, enabling efficient and cooperative target search.}
    \label{fig6}
\end{figure}

Efficient visual target navigation is crucial for autonomous robots in unfamiliar environments (figure~\ref{fig:sub2}), yet existing single-robot methods often lack commonsense reasoning and fail to scale to multi-robot settings. To address this, Yu \emph{et al.}\ propose \textbf{Co-NavGPT}\cite{yu2023co}, a framework that leverages a VLM as a global planner for cooperative multi-robot navigation. Co-NavGPT fuses sub-maps from multiple robots into a unified semantic representation, encoding robot states and frontier regions to guide coordinated exploration as depicted in figure \ref{fig6}. Using structured visual prompts, the VLM allocates frontier goals based on spatial and semantic cues, enabling efficient, zero-shot multi-robot planning without task-specific training. Experiments on the HM3D dataset show that it substantially improves success rate and navigation efficiency compared to baselines, with ablation studies underscoring the value of semantic priors. Real-world deployment on quadruped robots further validates its practicality and real-time performance. While limitations remain—particularly in multi-floor navigation and object detection robustness—Co-NavGPT demonstrates the potential of VLM-based reasoning for scalable, collaborative exploration in complex environments. Future work can focus on deeper integration of VLMs with embodied agents in 3D environments, emphasizing interactive decision-making, dynamic replanning, and real-time closed-loop control.

VLN agents often struggle during deployment due to distribution shifts and a lack of feedback, \textbf{A}ctive \textbf{TE}st-time \textbf{N}avigation \textbf{A}gent (\textbf{ATENA}) \cite{ko2025active} is a test-time active learning framework designed to improve VLN performance in unfamiliar environments. ATENA addresses by integrating \textbf{episodic HRI} and \textbf{self-guided learning to calibrate uncertainty}. The method introduces \textit{Mixture Entropy Optimization}, which blends action and pseudo-expert distributions to adjust confidence based on success or failure, and a \textit{Self-Active Learning strategy} that allows agents to evaluate outcomes based on prediction certainty. This dual approach fosters adaptive, well-grounded decision-making. Evaluations on REVERIE, R2R, and R2R-CE benchmarks show significant improvements over baselines. While promising for interactive tasks like VLN, the framework’s generalizability to less interactive settings remains an open question for future exploration.

Allgeuer \emph{et al.}\ \cite{allgeuer2024robots} present a modular framework that grounds an LLM in a robot’s sensory and motor capabilities to enable natural, open-ended \textbf{human–robot dialogue and collaboration}. Their system integrates modules for speech recognition and synthesis, object detection, pose estimation, and gesture recognition, with the LLM orchestrating these components via text-based prompts. Adding new skills or perception modules requires only updating the system prompt, no retraining, highlighting the flexibility and social intelligence afforded by this design. Future work could enhance spatial reasoning in object status updates.


\textbf{Robots Can Feel}\cite{lykov2024robots} introduces a novel ethical reasoning framework for robots that integrates logical inference with simulated human-like emotions to enable morally informed decision-making. Central to this approach is the \textbf{Emotion Weight Coefficient (EWC)}, a tunable parameter that modulates the influence of emotions in robot behavior, allowing flexible adaptation to different contexts and robot types. The framework operates independently of specific base models and was evaluated using eight SoTA LLMs, both commercial and open-source. Results indicate that varying the EWC consistently affects ethical decisions across models, as confirmed by ANOVA analysis. This emulates human ethical reasoning in robots, with potential applications in cognitive and social robotics where human-like behavior is advantageous. By adjusting emotional parameters, the framework can produce context-dependent responses, introducing variability similar to human moral reasoning. While such uncertainty may be unsuitable for high-stakes industrial tasks, it is valuable for domains like child interaction, domestic assistance, and human-behavior emulation.

Accurate spatial understanding from visual inputs is essential for robotic operation in unstructured environments, yet it remains an inherently ill-posed problem. Classical methods can estimate relative poses effectively but often lack data-driven priors to resolve ambiguities, a challenge amplified in multi-robot systems requiring frequent and precise localization. To address this, Blumenkamp \emph{et al.}\ present \textbf{CoViS-Net}\cite{blumenkamp2024covis}, a \textbf{Co}operative multi-robot \textbf{Vi}sual \textbf{S}patial foundation model that learns spatial priors from data. Unlike prior approaches evaluated offline, it is designed for online, real-time deployment on decentralized, platform-agnostic robots without reliance on external infrastructure. The model focuses on relative pose estimation and local \textit{bird’s-eye view} (BEV) prediction, accurately inferring poses without camera overlap and predicting BEVs for occluded regions. Validated on real-world indoor datasets and deployed for multi-robot formation control, CoViS-Net improves BEV prediction by 8.75\% over pose-agnostic aggregation methods and achieves precise trajectory tracking. This work highlights the potential of vision-only cooperative models for scalable, real-world multi-robot applications and opens avenues for integrating such models with downstream multi-agent control policies.

Current SoTA approaches to task anticipation primarily rely on data-driven deep learning models and LLMs, typically at a high level of abstraction and often requiring extensive training data. However, these methods have shown limitations in handling multistep, hierarchical decision-making that requires structured reasoning with domain knowledge. Arora \emph{et al.}\ \cite{arora2024anticipate} address this by employing the \textbf{Planning Domain Definition Language} (PDDL) and the \textbf{ Fast Downward} (FD) planner to generate detailed action plans for anticipated tasks. Their framework uses LLMs to predict high-level goals from partial observations, which are then treated as objectives in a classical planning pipeline. Future directions can include scaling to more complex domains, integrating probabilistic planning, deploying on physical robots, and personalizing task anticipation based on user preferences. Classical planning using the PDDL guarantees goal achievement through valid action sequences but struggles to represent temporal aspects such as concurrent actions without extensive domain modifications. Human experts can address this by decomposing goals into subgoals for parallel execution by multiple agents. While LLMs lack formal success guarantees, they can leverage commonsense reasoning to compose plausible action sequences. Bai \emph{et al.}\ introduce \textbf{TWOSTEP}\cite{bai2024twostep}, a framework that combines LLM-based goal decomposition with classical planning to approximate human multi-agent planning strategies. TWOSTEP assigns partially independent subgoals to multiple helper agents, with a main agent completing the remaining tasks. This approach yields significantly faster planning times and shorter execution lengths than direct multi-agent PDDL solutions, while retaining execution guarantees. Furthermore, the subgoals inferred by LLMs closely align with those produced by human experts, demonstrating the framework’s effectiveness across both symbolic and embodied domains.

The recent success of LLMs such as ChatGPT and GPT-4 has demonstrated their versatility across a wide range of tasks. However, their potential for multi-agent planning remains underexplored. This domain presents unique challenges by combining complex agent coordination with planning, making it difficult to leverage external reasoning tools effectively. Chen \emph{et al.}\ \cite{chen2024solving} investigates the use of LLMs for \textbf{multi-agent path finding (MAPF)}, also known as multi-robot route planning. Initial experiments show that LLMs can generate valid plans in simple, obstacle-free environments with few agents but fail on more complex benchmark maps. The authors analyze these failures, attributing them to limitations in context length, obstacle representation, and planning ability. Through extensive experiments, they provide evidence supporting this analysis and outline how interdisciplinary approaches could help address these challenges in future models.

Integrating language understanding into robotic systems enables advanced spatial task execution but introduces unique challenges, particularly in pattern formation. \textbf{ZeroCAP}\cite{venkatesh2025zerocap} (Zero-Shot multi-robot Context Aware Pattern) is a zero-shot, context-aware pattern formation framework that combines LLMs with MRS. It leverages natural language instructions to generate precise spatial configurations, integrating VLMs, segmentation, and shape descriptors to translate linguistic input into actionable robot placements. For example, when instructed to “surround the incorrectly parked car at the corners,” the system identifies the target object and autonomously positions robots to form the required pattern. By decoupling spatial reasoning from visual processing and employing edge–vertex representations, ZeroCAP overcomes limitations of 2D-trained VLMs, achieving accurate pattern formation in diverse scenarios. Extensive experiments demonstrate its effectiveness in tasks such as surrounding, caging, and infilling, highlighting its adaptability and potential for applications in areas like surveillance and logistics. Although currently constrained to 2D environments, the framework is designed for future extensions to dynamic and 3D settings.

Effective collaboration is essential for teams of autonomous robots to navigate large, unknown environments. \textbf{SayCoNav}\cite{rajvanshi2025sayconav} is a novel framework that employs LLMs to automatically generate and adapt collaboration strategies for heterogeneous multi-robot teams. Each robot uses LLM-based global and local planners to devise decentralized action plans, which are continuously refined through inter-robot communication. They used the ProcTHOR framework\cite{deitke2022️} with the AI2-THOR simulator\cite{kolve2017ai2} . Evaluated on Multi-Object Navigation (MultiON) tasks in procedurally generated realistic environments, SayCoNav enables robots to exploit their complementary capabilities, improving search efficiency by up to 44.28\% compared to baselines. The framework demonstrates strong adaptability to dynamic conditions and varied team compositions. While SayCoNav effectively coordinates heterogeneous agents, challenges remain in handling tasks that require tightly coupled manipulation and perception, as well as mitigating occasional LLM hallucinations. Future work includes deploying SayCoNav on real multi-robot platforms to further validate its robustness.

Embodied agents powered by LLMs often struggle with collaborative tasks due to limited communication strategies and inefficient task allocation, leading to incoherent actions and execution errors. To address these challenges, Zu \emph{et al.}\ propose \textbf{Cooperative Tree Search (CoTS)}\cite{zu2025collaborative}, a framework that integrates LLMs with a modified \textit{Monte Carlo Tree Search (MCTS)} to enhance multi-agent planning and coordination. CoTS enables agents to deliberate over multiple strategic plans within a collaborative search tree guided by LLM-driven rewards, while a plan evaluation module ensures stability by updating strategies only when necessary. Unlike prior methods such as CoELA\cite{zhang2023building} and RoCo\cite{mandi2024roco} , which rely on either local decisions or single-round dialogues, CoTS supports structured, long-term collaboration and informed decision-making. Experiments on complex embodied environments demonstrate that CoTS significantly improves planning efficiency, communication quality, and task success. Although its performance depends on the underlying LLM, with GPT-4 outperforming GPT-3.5-turbo, CoTS consistently enhances agent collaboration through structured planning search.

Chen \emph{et al.}\ \cite{chen2023multi} investigates consensus seeking in multi-agent systems driven by LLMs, a fundamental challenge in collaborative decision-making. In this setting, each agent holds a numerical state and negotiates with others to converge on a common value. Experiments reveal that, without explicit instructions, LLM-driven agents predominantly adopt an averaging strategy to achieve consensus, with occasional use of alternative methods. The study further examines how factors such as agent number, personality traits, and network topology influence the negotiation dynamics. Increasing the number of agents is found to mitigate hallucinations and stabilize decision-making. These insights offer a foundation for understanding LLM-driven behaviors in more complex multi-agent scenarios. Additionally, consensus seeking is applied to a multi-robot aggregation task, demonstrating the potential of LLMs for zero-shot collaborative planning. However, the current approach is limited by its reliance on simple numerical states, a single LLM (GPT-3.5), and a low planner update rate, which constrains its performance in high-speed multi-robot applications.

\section{LLMs and Reasoning in VLN}
\label{section_6}

Li \emph{et al.}\ \cite{li2025large} present the first comprehensive survey of integrating LLMs into MRS, categorizing their roles in task allocation, motion planning, action generation, and human–robot interaction. They showcase applications across domains, from household robotics to formation control and target tracking, while identifying challenges such as reasoning limitations, hallucinations, latency, and benchmarking gaps. Finally, they outline research directions in fine-tuning, robust evaluation, and task-specific model design to accelerate the deployment of LLM-powered MRS in real-world scenarios. Whereas Wang \emph{et al.}\ \cite{wang2024large} survey the integration of LLMs and multimodal LLMs into robotic task planning, highlighting their advanced reasoning and instruction understanding capabilities. They introduce a GPT-4V-based framework that fuses language instructions with visual inputs to generate detailed action plans, demonstrating its efficacy across nine embodied task datasets. While these results affirm the promise of multimodal LLMs as “robotic brains,” challenges remain in model interpretability, robustness, safety, and sim-to-real transfer. Addressing these issues through standardized evaluation, adversarial training, policy adaptation, and ethical oversight will be crucial as we move toward fully simulated development and deployment of intelligent robotic systems.

Yu \emph{et al.}\ \cite{yu2023l3mvn} present \textbf{L3MVN}, a novel framework for visual target navigation that leverages LLMs to inject common-sense knowledge into object search tasks. Unlike traditional methods that require extensive training, their approach introduces zero-shot and feed-forward paradigms to identify semantically relevant frontiers from a map as long-term goals. Experiments on Gibson and HM3D show that L3MVN achieves good SR and generalization. Ablation studies confirm that language-driven semantic reasoning enhances exploration efficiency. Real-world tests validate its practical applicability, highlighting LLMs’ potential in robotics for efficient, generalizable navigation without costly training. Similarly, \textbf{Language-guided exploration (LGX)} \cite{dorbala2023can} is a zero‑shot object‑goal navigation algorithm that combines LLMs for sequential decision‑making with open‑vocabulary vision–language grounding for target detection. On RoboTHOR, LGX performs well in SR, and the impact of different LLM prompting strategies and validating LGX’s real‑world efficacy was analyzed, highlighting its strong language‑guided exploration capabilities.

Hong \emph{et al.}\ introduce $Ego^2$-Map \cite{hong2023learning}, a contrastive learning method that aligns egocentric views with top-down semantic maps to embed spatial and object relationships into visual representations. Using a ViT encoder trained on HM3D, $Ego^2$-Map transfers map-derived semantics, such as object layouts and connectivity, into the agent’s first-person features. Although $Ego^2$-Map requires semantic maps or annotations for training, its strong generalization and improved planning capabilities suggest a promising direction for map-aware visual learning in navigation. Whereas Chen \emph{et al.}\ \cite{chen2022weakly} introduce a \textbf{multi-granularity map} for VLN that encodes both fine-grained object details (e.g., color, texture) and semantic classes. To refine this representation, they add a weakly supervised auxiliary task that trains the agent to pinpoint instruction-relevant objects on the map. The enriched map and parsed instructions feed into a waypoint predictor, yielding a 4\% absolute boost in SR on the VLN-CE dataset. While effective, their approach relies on ground-truth semantics and 2D top-down maps (limiting multi-floor navigation); future work can explore 3D mapping, commonsense grounding, and real-world deployment.

Embodied VLN demands integrated understanding, perception, and planning, yet current models, even GPT-4  rely on single‑round self‑reasoning and struggle with complex tasks. \textbf{DiscussNav} \cite{long2024discuss} is a zero‑shot framework that treats specialized LLMs as \textit{domain experts}. It queries these experts on instruction interpretation, scene perception, and progress estimation before acting. On R2R, this multi‑expert discussion outperforms the top zero‑shot baseline across all metrics, and real‑robot trials confirm its advantage over single‑round reasoning. Whereas Zhu \emph{et al.}\ propose \cite{zhu2020vision} Auxiliary Reasoning Navigation (\textbf{AuxRN)}) to enhance VLN by leveraging semantic information often overlooked in prior work. AuxRN introduces four self-supervised auxiliary tasks: explaining past actions, estimating navigation progress, predicting future orientation, and evaluating trajectory consistency, to guide the agent in learning richer semantic representations. These tasks provide additional training signals that improve both task performance and generalization. Experiments show that AuxRN significantly boosts navigation accuracy, offering a promising direction for incorporating common-sense reasoning in future VLN models.

Some of the recent frameworks couple prompt‑engineering guidelines with a high‑level function library, enabling ChatGPT to tackle diverse robotics tasks, simulators, and platforms. Results demonstrate ChatGPT’s effectiveness in executing complex robotics workflows using natural language. \textbf{PromptCraft} \cite{vemprala2024chatgpt} is an open‑source tool featuring a collaborative repository of optimized prompts and a sample simulator integrated with ChatGPT, streamlining adoption of conversational AI in robotics research. Whereas Mower \emph{et al.}\ \cite{mower2024ros} propose a user-friendly framework for robot programming that enables non-experts to instruct robots using natural language prompts and contextual information from the Robot Operating System (ROS). The system integrates LLMs with ROS, allowing task instructions via a chat interface. It features automatic behavior extraction from LLM outputs, execution of ROS actions, and supports sequence, behavior tree, and state machine modes. Additionally, it incorporates imitation learning to expand the action library and uses feedback from humans and the environment to refine LLM outputs. Experiments demonstrate the framework’s robustness, scalability, and effectiveness across varied tasks, including long-horizon operations, tabletop manipulation, and remote supervision.

Lisondra \emph{et al.}\ introduce \textbf{Adaptive Text Dreamer (ATD)} \cite{zhang2025cross}, a dual-branch VLN framework that leverages LLMs for efficient, language-based imagination under partial observability. Mimicking human cognition, ATD features a left-brain module for logical reasoning and a right-brain module for semantic imagination, both using fine-tuned Q-formers to dynamically activate domain knowledge. A cross-interaction mechanism integrates imagined semantics into a navigation policy, enhancing decision-making. Evaluated on the R2R benchmark, ATD outperforms prior methods with fewer parameters and reduced computational cost, demonstrating the effectiveness of linguistic abstraction for guided imagination in embodied navigation. Whereas \textbf{VISTA} \cite{huang2025vista} introduces a novel imagine-and-align strategy for VLN, addressing limitations in long-horizon tasks faced by conventional observe-and-act models. It leverages a diffusion model to generate visual goal imaginations conditioned on language and local observations, refined through a perceptual alignment module. An adaptive scheduler dynamically balances static and dynamic goal prediction, enhancing reasoning and decision-making. VISTA achieves good results on R2R \cite{anderson2018vision} and RoboTHOR \cite{deitke2020robothor}, notably improving navigation in ambiguous settings. Limitations include generative fidelity, computational overhead, and reliance on hand-tuned parameters. Future work can aim to enhance real-world transferability through efficient generation and adaptive scheduling.

Dual Object Perception-Enhancement (\textbf{DOPE}) Network enhances VLN by refining language understanding and cross-modal object reasoning \cite{yu2025dope}. A \textit{Text Semantic Extraction} (TSE) module identifies key phrases, which \textit{Text Object Perception-Augmentation} \textbf{(TOPA)} then uses to enrich instruction details. Simultaneously, \textit{Image Object Perception-Augmentation} \textbf{(IOPA)} models latent object relationships between vision and language. Evaluated on R2R and REVERIE, it outperforms prior methods, demonstrating improved navigation accuracy and robustness. Sililarly, \textbf{GroundingMate} \cite{liu2025groundingmate} is a plug-and-play, model-agnostic method to address the overlooked object grounding challenge in Goal-Oriented VLN. While prior work emphasizes navigation success, this focuses on accurately identifying target objects at the destination. It employs a \textit{confusion detection} mechanism to determine when the agent struggles with object localization and then invokes a \textit{Multi-Modal Large Language Model (MLLM)} for assistance. The agent first extracts relevant object details via an LLM and then performs multi-stage evaluation using the MLLM to refine predictions. Without requiring retraining, the method integrates with existing VLN models and shows significant improvements on REVERIE and SOON, demonstrating its effectiveness and broad applicability.

While earlier work has focused on single-robot setups with single-threaded LLM planning, recent efforts are advancing toward more scalable, interactive frameworks. Mandi \emph{et al.}\ \cite{mandi2024roco} introduce an approach to multi-\underline{ro}bot \underline{co}llaboration \textbf{(RoCo)} by leveraging pre-trained LLMs for both high-level dialogue-based coordination and low-level path planning. Robots engage in natural language discussions to reason about task strategies, decompose goals into sub-tasks, and generate waypoints, which are refined using environmental feedback such as collision checks. Real-world experiments highlight the system’s flexibility, including seamless human-in-the-loop collaboration. While performance remains below perfect accuracy, this work lays a strong foundation for developing more capable and interpretable multi-agent systems guided by language. Whereas Chen \emph{et al.}\ introduce \textbf{AO-Planner} \cite{chen2025affordances}, an affordance-oriented framework for continuous VLN that addresses the gap between high-level LLM-based planning and low-level motion control. Unlike prior zero-shot approaches limited to abstract graph navigation, this leverages \textit{visual affordance prompting (VAP)} to enable LLMs to make grounded motion decisions. It segments navigable regions using SAM\cite{ren2024grounded} and prompts the LLM to select waypoints and generate low-level paths. A high-level module, \textit{PathAgent}, converts pixel-level plans into 3D coordinates for execution. Despite some limitations in the underlying foundation models (e.g., Grounding DINO inaccuracies), this work marks a key step in connecting LLMs to real-world navigation by enabling pixel-to-3D motion planning. Future work can explore integrating LLMs with learned waypoint predictors for broader generalization beyond simulators.

\textbf{NavGPT} \cite{zhou2023explicit} is a purely LLM-driven agent that, without additional training, predicts sequential actions by combining textual inputs, scene descriptions, navigation history, and prospective directions. While NavGPT’s zero-shot performance is constrained by the richness of visual-text mappings and object tracking, GPT-4’s reasoning traces highlight substantial promise. As future work, fine-tuning VLMs specifically for navigation promises even greater robustness and versatility. Integrating multimodal inputs, high-level planning modules, and collaboration with specialized downstream models could unlock versatile VLN agents. Whereas \textbf{NavGPT-2} \cite{zhou2024navgpt} addresses the performance gap between VLN specialists and LLM-based navigators by tightly integrating frozen LLMs with navigation policy networks. They align visual inputs within the LLM to leverage its rich language reasoning and then feed its latent representations into a downstream policy for action prediction. This fusion retains the LLM’s interpretive strengths, enabling natural language reasoning during navigation while matching specialist models in efficiency and accuracy. Their experiments demonstrate data-efficient learning and seamless vision–language–action alignment, paving the way for versatile agents that understand and execute free-form human instructions. 

Kim \emph{et al.}\ \cite{kim2024survey} present a comprehensive survey on the transformative role of LLMs in robotics, focusing on their integration into core components, communication, perception, planning, and control. Centered on models developed post-GPT-3.5, the study emphasizes text-based and emerging multimodal applications. It outlines how LLMs address limitations of traditional methods, offering structured guidance on prompt engineering and practical examples to support integration. \textbf{TrustNavGPT}\cite{sun2024trustnavgpt} is another LLM-based navigation agent that leverages affective cues, such as tone and inflection, to gauge the trustworthiness of spoken instructions and improve decision safety. By integrating audio-driven uncertainty modeling with text understanding, it achieves a 70.5\% SR in detecting ambiguous commands and an 80\% target-finding rate. It also demonstrates 22\% greater resilience to adversarial audio perturbations. While audio processing adds computational overhead and depends on input quality, future work may explore denoising and retrieval-augmented strategies to boost efficiency and robustness. This approach paves the way for more reliable, audio-directed robotic navigation.

As we know, recent LLM-based task planners have shown strong performance but are typically limited to simple tasks involving homogeneous robots. Addressing the demands of complex, long-horizon tasks requiring coordination among heterogeneous robots, \textbf{COHERENT}\cite{liu2024coherent}, an LLM-driven framework for multi-robot collaboration involving quadrotors, robotic arms, and robot dogs. The framework employs a \textit{Proposal-Execution-Feedback-Adjustment (PEFA)} loop, where a centralized planner decomposes tasks into subtasks, assigns them to individual robots, and iteratively refines the plan based on each robot’s self-reflective feedback. Results demonstrate that COHERENT significantly outperforms prior methods in both success rate and efficiency. While LLMs have shown potential in enhancing reasoning and interpretability for VLN, their offline usage often leads to a domain gap due to misalignment with real-world navigation tasks. To address this, \textbf{Navigational Chain-of-Thought (NavCoT)}\cite{lin2025navcot}, a parameter-efficient, in-domain training strategy enabling LLMs to make self-guided navigational decisions. At each step, NavCoT prompts the LLM to: (1) imagine the next observation based on the instruction, (2) align it with candidate views, and (3) decide the action via step-wise reasoning. This disentangled reasoning simplifies action prediction and enhances decision-making. Experiments on benchmarks like R2R, RxR, and R4R show that NavCoT significantly outperforms direct action prediction methods, achieving a ~7\% improvement on R2R over a recent GPT-4 baseline. NavCoT demonstrates the potential for scalable, task-adaptive LLM-based navigation in real-world robotics.

While GPT-4 \cite{chen2023open} based zero-shot agents have shown impressive language understanding on the R2R benchmark, they often struggle with obstacle avoidance and richer instruction sets. \textbf{CorNav} \cite{liang2024cornav} is a zero-shot, LLM-powered framework that continually refines its plan using real-time environmental feedback and employs specialized \textit{“domain experts”} for instruction parsing, scene interpretation, and action refinement. Accompanied by a high-fidelity \textit{Unreal Engine 5}\footnotemark simulator and the \textit{NavBench} benchmark, CorNav demonstrates strong adaptability across multiple zero-shot tasks. Whereas Wang \emph{et al.}\ \cite{wang2024navigating} introduce a novel evaluation framework for VLN that systematically diagnoses model performance across fine-grained instruction types. Grounded in a context-free grammar (CFG) of navigation tasks, the framework uses a semi-automated CFG construction process with LLMs to generate data across five core instruction categories: \textit{direction change, landmark recognition, region recognition, vertical movement, and numerical comprehension}. Evaluations on their \textit{NAVNUANCES} benchmark reveal model-specific limitations, including poor numerical reasoning and directional biases. Notably, a zero-shot agent enhanced with GPT-4, vision demonstrates improved landmark recognition, highlighting the value of strong vision-language alignment in advancing VLN capabilities.

\footnotetext{https://www.unrealengine.com/}

VLN-CE enables agents to follow human instructions in realistic settings but often suffers from limited world knowledge and inadequate obstacle avoidance. To tackle that, recently \textbf{RAGNav} \cite{bao2025enhancing}, which builds a navigation knowledge base and uses Retrieval-Augmented Generation \textbf{(RAG)} \cite{lewis2020retrieval} to enrich LLM inputs for more accurate route planning. But the existing data augmentation methods often produce overly detailed, step-wise instructions that fail to reflect natural user communication, and they neglect global scene context. To overcome these limitations, Wang et al. propose \textbf{NavRAG}\cite{wang2025navrag}, a RAG framework that produces diverse, user-style instructions for VLN. NavRAG constructs a hierarchical scene description tree using LLMs to capture both global layout and local details, simulates varied user demands, and generates realistic instructions via retrieval and generation.

\section{What if the Ambiguity exists in the language instructions?}
\label{sec:ambiguity}

A major challenge in VLN is navigating effectively under uncertainty caused by ambiguous instructions and limited environmental observations. Inspired by human behavior, \cite{wang2020active} work equips agents with active information-gathering capabilities to improve decision-making. The proposed end-to-end framework learns an exploration policy that determines (i) when and where to explore, (ii) which information to prioritize, and (iii) how to refine navigation based on gathered data. Unlike prior methods, it directly addresses ambiguity and partial observability through an \textit{exploration module} that enhances robustness and significantly boosts navigation performance. Similarly, Embodied Learning-By-Asking (\textbf{ELBA}) \cite{shen2025elba} is another framework that enables agents to actively ask questions to resolve ambiguities during navigation and task execution, rather than passively following instructions. Evaluated on the \textit{TEACh} \cite{padmakumar2022teach} dataset, ELBA learns when and what to ask, resulting in improved task performance over baselines lacking question-asking abilities. This work highlights the importance of interactive question-asking for enhancing agent autonomy in complex, real-world environments.

Reflecting real-world conditions, the agent may lack full navigation knowledge and can request guidance through subgoal instructions when lost. To enable this, Nguyen \emph{et al.}\ propose Imitation Learning with Indirect Intervention\textbf{ (I3L)}, a framework for incorporating language-based assistance. Similarly, \textbf{VNLA} \cite{nguyen2019vision} (Vision-based Navigation with Language Assistance) involves an agent navigating realistic indoor settings to find objects by following high-level language instructions. In contrast, \textbf{CoNav} \cite{hao2025conav} is a cross-modal reasoning framework where a pretrained 3D text model provides structured spatial-semantic information to guide an image-text navigation agent. At the core of CoNav is \textit{Cross-Modal Belief Alignment}, which transmits textual predictions from the 3D text model to help \textbf{resolve ambiguity in navigation}. By fine-tuning on a compact 2D–3D–text dataset, the agent effectively integrates visual and spatial-semantic signals. CoNav surpasses existing approaches across four embodied navigation benchmarks (R2R, CVDN, REVERIE, SOON) and two spatial reasoning challenges (ScanQA \cite{azuma2022scanqa}, SQA3D \cite{ma2022sqa3d}), often generating shorter and more efficient routes. This study demonstrates the promise of leveraging 3D text-based reasoning for more reliable and practical embodied navigation solutions.

Traditional VLN assumes that language commands are always feasible within a given environment. However, as we know, real-world tasks often involve ambiguous instructions or dynamic environments where commands may not be executable. Mobile App Tasks with Iterative Feedback (\textbf{MoTIF}) \cite{burns2022dataset} is the VLN dataset to explicitly model task uncertainty, offering greater linguistic and visual diversity than previous benchmarks. It enables the study of feasibility prediction and supports more realistic evaluations of VLN models. The authors assess prior methods on this dataset and demonstrate the need for more robust vision-language approaches. The future directions can include learning hierarchical representations through tools like Screen2Vec \cite{li2021screen2vec}, icon embeddings, and leveraging app view hierarchies using Transformers or Graph Neural Networks to better model structured features and app affordances. So to enable robots to function in human environments, they must understand and execute natural language instructions while resolving ambiguities through dialogue. To support this, Padmakumar \emph{et al.}\ introduce \textbf{TEACh} \cite{padmakumar2022teach}, a dataset of over 3,000 human-human dialogues involving household tasks in simulation. A \textit{Commander}, with task knowledge, guides a \textit{Follower} who interacts with the environment and asks clarifying questions to complete tasks like \textit{MAKE COFFEE} or \textit{PREPARE BREAKFAST}. They propose benchmarks to evaluate models on dialogue understanding, language grounding, and task execution.

\begin{figure}[htbp]
    \centering
    \Description{Key components of future directions.}
    \includegraphics[width=0.5\linewidth]{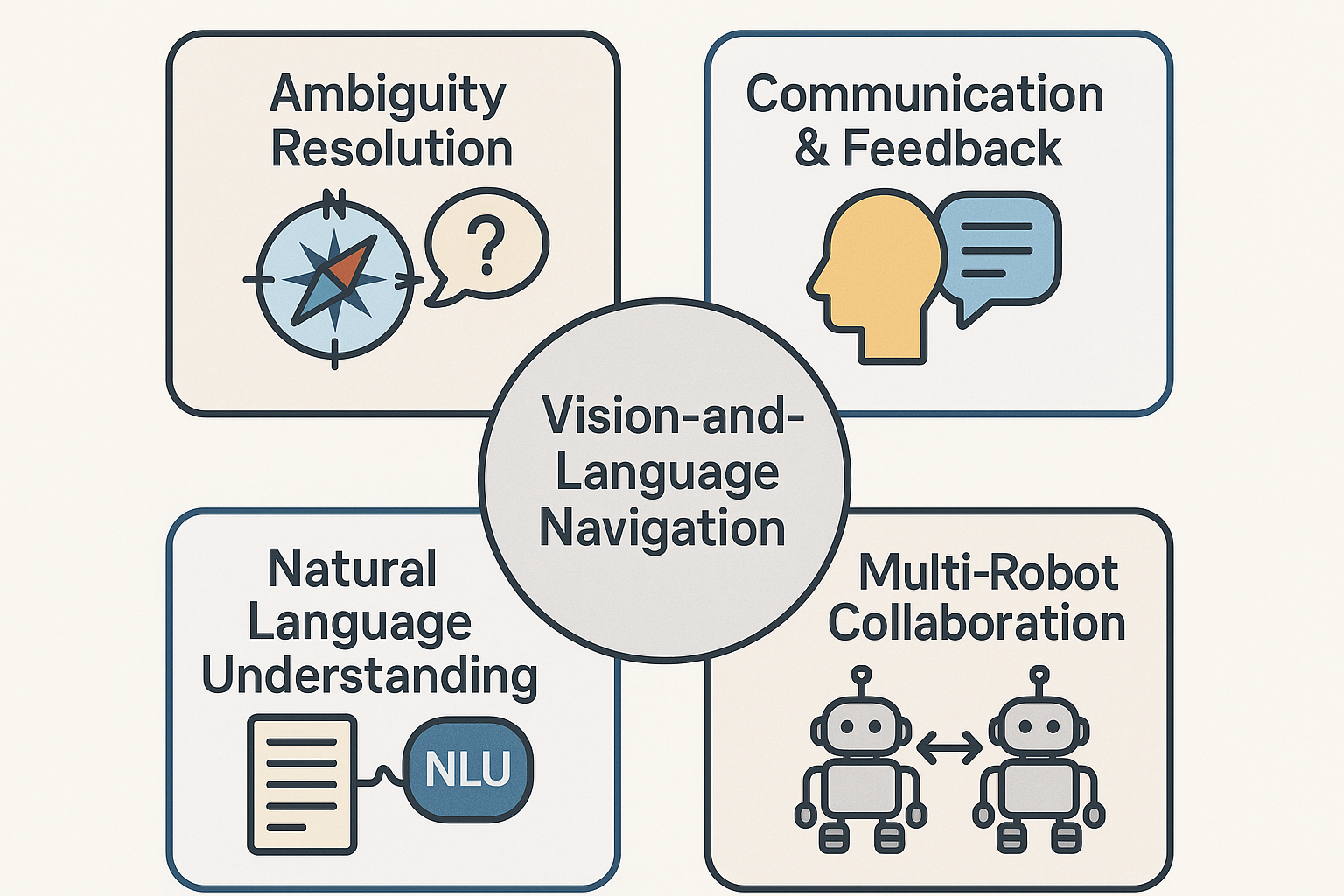}
    \captionsetup{justification=justified}
    \caption{\small Some of the key components of future directions include ambiguity resolution, communication and feedback, and decentralized decision-making with dynamic role assignment.}
    \label{fig7}
\end{figure}

\section{Conclusion and Future Directions}
\label{Future}

In this paper, we systematically introduced the prior work, then the growth and diversification and the recent developments in the emerging field of VLN. We broadly review the VLN methodologies and classified current solutions considering the MRS and HRI in VLN. The work also considers the recent applications of LLMs in Multi-agent systems. With that comprehensive review of SoTA methods, we now consolidate the remaining open challenges and outline promising avenues for future research and some of them are depicted in the Fig.~\ref{fig7}.

\subsection{Limited HRI Capabilities}
\label{future_1}

As we have seen in the earlier sections, some work has been done in the domain of improving the HRI, but still, the effective human–robot collaboration is often limited by unidirectional communication, especially in high-stakes or uncertain situations. For example, in the healthcare settings, caregivers may struggle to convey nuanced preferences to assistive robots, leading to task errors. Future research should prioritize \textbf{bidirectional interaction}, equipping robots with dialogue systems that both interpret human intent and seek clarifications. By integrating real-time voice commands and feedback loops, robots can proactively confirm instructions and provide ongoing status updates, fostering more reliable and collaborative teamwork.  

\subsection{Ambiguous Instructions in Navigation Tasks}
\label{future_2}

\begin{figure}[htbp]
    \centering
    \Description{Artist impression of VLN in MRS.}
    \includegraphics[width=0.6\linewidth]{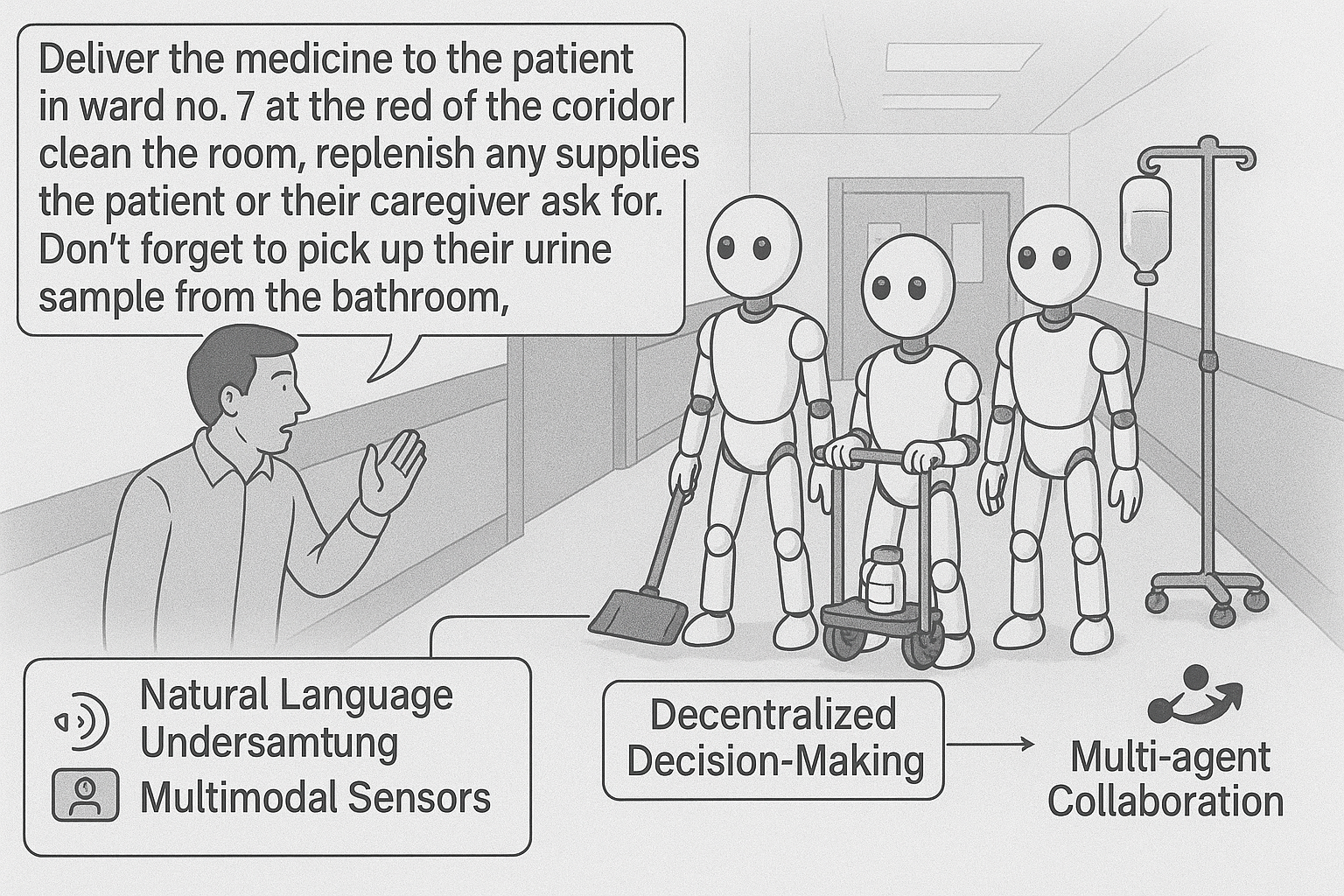}
    \captionsetup{justification=justified}
    \caption{\small Artist impression of VLN based MRS in healthcare settings: Robots interpret human instructions via NLU, utilize multi-modal sensory input, ambiguity resolution and collaborate through real-time feedback.}
    \label{fig13}
\end{figure}

As we have seen in the section ~\ref{sec:ambiguity}, ambiguous and/or incomplete instructions (we can observe the word "red" instead of the word "end" in the instruction as depicted in the Fig.\ref{fig13}) can significantly impair navigation accuracy, as current models often fail to interpret vague commands when multiple candidates exist. Future work should enhance NLU by incorporating \textbf{contextual reasoning} and \textbf{interactive clarification}. By enabling robots to ask targeted questions and confirm user intent, such systems can resolve uncertainties and achieve more reliable, human-like navigation. In other words, key components include ambiguity resolution, contextual reasoning, and decentralized decision-making with dynamic role assignment. These capabilities support scalable and efficient robot coordination in real-world environments.

\subsection{Lack of Robust Multi-Robot Coordination}
\label{future_3}
Coordinating task allocation among multiple robots grows increasingly complex as team size expands, yet most VLN research remains single-agent. This gap hinders consensus-building, conflict resolution, and resource optimization. Consider, for example, warehouse robots that collide or duplicate work due to poor communication. Future work should explore \textbf{VLN-driven, decentralized frameworks} for dynamic role assignment, enabling robots to share intentions, distribute tasks efficiently, and resolve conflicts collaboratively.

\subsection{Sim-to-Real Transfer Challenge}
\label{future_4}
Simulation-trained VLN models often underperform in real-world settings due to unmodeled dynamics such as lighting changes, moving objects, and sensor noise. While initial efforts have sought to bridge this gap, more remains to be done. Integrating noise-augmented simulations with real-world sensor data, varying illumination, motion, and environmental noise can yield hybrid frameworks that improve VLN adaptability and generalization across diverse environments.

\section*{APPENDIX}
We have summarized some of the key approaches in VLN, considering single and multi-agents, in Table~\ref {SOTA_class}.
\small
\begin{longtable}{|>{\arraybackslash}p{2cm}
                  |>{\arraybackslash}p{4.5cm}
                  |>{\arraybackslash}p{2.5cm}
                  |>{\arraybackslash}p{3cm}
                  |>{\arraybackslash}p{3cm}|}
\caption{Summary of Representative SoTA VLN Research Contributions. \label{SOTA_class}}\\
\hline
\rowcolor{gray!20}
\textbf{Reference} & 
\textbf{Description/Contribution} & 
\textbf{Evaluation Metrics used} & 
\textbf{Dataset and Simulator} & 
\textbf{Drawbacks or Limitations} \\ 
\hline
\endfirsthead

\multicolumn{5}{c}%
{{\small \tablename\ \thetable{} -  Continued from previous page}} \\
\hline

\hline
\rowcolor{gray!20}
\textbf{Reference} & 
\textbf{Description/Contribution} & 
\textbf{Evaluation Metrics used} & 
\textbf{Dataset and Simulator} & 
\textbf{Drawbacks or Limitations} \\  
\hline
\endhead

\hline \multicolumn{5}{|r|}{{\small Continued on next page}} \\ \hline
\endfoot


\endlastfoot
        
     Audio Visual Language Maps (AVLMaps) for Robot Navigation \cite{huang2023audio} & 1) Integrates visual and sound semantics into a unified map. 2) Enables robots to navigate to the goals specified by goal images or natural language (e.g., "go to the sound of baby crying"). 3) Multimodal prompts, such as “go to the \{image of a table\} where the sound of the microwave was heard”.
     & Recall and Average min. distance. & \textbf{MP3D} dataset and \textbf{Habitat} simulator. & 1) Sensitive to the noise in the recorded audio, and it assumes a static environment throughout their lifetime. \\ 
     \hline
     Instance-Level Semantic (SI) Maps for Vision Language Navigation \cite{nanwani2023instance} & 1) A memory-efficient mechanism for creating a semantic spatial representation of the environment, which is directly applicable to robots navigating in real-world scenes. 2) Allows indoor embodied agents to perform complex instance-specific goal navigation in object-rich environments.& SR. & \textbf{MP3D} dataset in the \textbf{Habitat} simulator.
      & ~  \\ 
      \hline
     LANA: A Language-Capable Navigator for Instruction Following and Generation \cite{wang2023lana} & 1) LANA formalises human-to-robot and robot-to-human communication, conveyed using navigation-oriented natural language, in a unified framework. & SR, SPL, CLS, nDTW, and SDTW. For REVERIE, RGS, and RGSPL. For Instruction generation: \textbf{SPICE} \cite{anderson2016spice}. & The agent is developed in virtual simulated environments. & 1) If the algorithm is deployed on a real robot in a real dynamic environment, the collisions during navigation can potentially cause damage to persons and assets. 2) LANA sometimes wrongly recognizes the storeroom as the bedroom.  \\ 
     \hline
     Visual Language Maps (VLMaps) for Robot Navigation \cite{huang2023visual} & 1) VLMaps support spatial language-based indexing that extends beyond specific object targets, allowing the creation of obstacle maps with open-vocabulary descriptions. 2) This work shows that VLMs can be leveraged to construct scene representations that are searchable, enabling LLMs to facilitate robot planning in environments containing unfamiliar objects and locations. & SR and SPL & \textbf{Habitat}, \textbf{AI2THOR} simulator and \textbf{MP3D} dataset. & 1) The approach is vulnerable to errors caused by 3D reconstruction noise and odometry drift during navigation. 2) It struggles to disambiguate objects when indexing landmarks in cluttered environments with visually similar items. \\ 
     \hline
     Iterative Vision-and-Language Navigation (IVLN) \cite{krantz2023iterative} & 1) IVLN is a paradigm for evaluating language-guided agents navigating in a persistent environment over time. 2) An agent follows an ordered sequence of language instructions that conduct a tour of an indoor space. & TL, NE, OS, nDTW, SR, and SPL. & Explore both a discrete VLN setting based on R2R episodes and navigation graphs (IR2R) and a continuous simulation VLN-CE setting (IR2R-CE). & 1) Limited to English instructions. 2) Deployed assistive robots should respond to more than English, and should be able to navigate cluttered, realistic home environments.  \\ 
     \hline
     ESC: Exploration with Soft Commonsense Constraints for Zero-shot Object Navigation \cite{zhou2023esc} & 1) Generalize to unseen environments and novel object types (Focuses on efficiently finding a goal object in unseen environments). 2) Transfer the commonsense knowledge in LLM into the object goal navigation task in a zero-shot manner. & SR and SPL. & MP3D, HM3D, and RoboTHOR \cite{deitke2020robothor}. & ~  \\ 
     \hline
     FM-Loc: Using Foundation Models for Improved Vision-based Localization \cite{mirjalili2023fm} & 1) Employ foundation models for both object detection and scene classification. 2) Perform both tasks in a zero-shot fashion and further grants greater flexibility on landmark and environment labels. 3) The capability to easily add objects to the vocabulary is one of the strengths of this method. & \textbf{Average translation error} between the query image camera poses and the poses of the retrieved reference images, and the \textbf{GT room labels} to calculate the percentage of correct room detections.  & Two datasets, each containing a reference and a query image set on two different floors of an office building. The first contains 111 images for query and 226 for reference, while the second consists of 101 images for both query and reference sets.  & In the hallway, the approach is behind the baselines due to the lack of distinct objects in that area. \\ 
     \hline
     Co-NavGPT: Multi-Robot Cooperative Visual Semantic Navigation using LLMs \cite{yu2023co} & 1) Employs LLMs to craft an efficient exploration and search policy for multirobot collaboration. 2) LLMs act as a \textbf{global planner}, assigning unexplored frontiers to each robot. & SR, SPL, and Distance to Goal (DTG) for multi-robot tasks. & HM3D\text{\_}v0.2 dataset & The SPL of the Random Sample Method (Baseline) is marginally higher than the authors' due to its superior continuous exploration from distant goals. \\ 
     \hline
    AerialVLN: Vision-and-Language Navigation for UAVs \cite{liu2023aerialvln} & 1) It is a city-level open environment dataset for aerial vision-and-language instruction-based navigation. 2) Combine the Cross-modal matching (CMA) model and look-ahead guidance (LAG). & SR, OSR, NE and SDTW & Simulator is developed based on \textbf{AirSim} and \textbf{Unreal Engine 4}. The AerialVLN dataset consists of 8,446 flying paths. & 1) Some results suggest that the agent has passed the goal location and failed to stop around it. 2) CMA could follow instructions at an early stage, but they cannot get back on track once they deviate. \\ 
    \hline
    Anticipate \& Act: Integrating LLMs and Classical Planning for Efficient Task Execution in Household Environments \cite{arora2024anticipate} & 1) Use the \textbf{P}lanning \textbf{D}omain \textbf{D}efinition \textbf{L}anguage (\textbf{PDDL}) as the action language, and use the Fast Downward (FD) solver to generate fine-granularity plans for any given task. 2) Will anticipate upcoming tasks and compute an action sequence that jointly achieves. & Four task anticipation performance measures: Miss Ratio, Partial Ordering Count (POC), Kendall rank correlation coefficient (KRCC), and Success Ratio & \textbf{VirtualHome}\cite{puig2018virtualhome}, a realistic simulation environment. & Some actions in this domain are irreversible, e.g., we cannot put the pieces of a cut fruit back together. \\ 
    \hline
    \textbf{H}ierarchical \textbf{O}pen-\textbf{V}ocabulary 3D \textbf{S}cene \textbf{G}raphs for Language-Grounded Robot Navigation \cite{werby2024hierarchical} & 1) \textbf{HOV-SG} is a hierarchical open-vocabulary 3D scene graphs representation for robot navigation. 2) The semantic decomposition of environments into \textbf{floors, rooms, and objects} performs long-horizon navigation across a multi-story environment in the real world. & mIOU (Mean Intersection over Union), F-mIoU (F-score Mean Intersection over Union,  and mAcc (Mean Accuracy). &  Replica \cite{straub2019replica} and ScanNet \cite{dai2017scannet} dataset and HM3D Semantics dataset \cite{yadav2023habitat}. & 1) The construction process is time-consuming, rendering the method unsuitable for real-time mapping. 2) Assumes a static environment and thus cannot handle dynamic environments.  \\ 
    \hline
    Think, Act, and Ask: Open-World Interactive Personalized Robot Navigation \cite{dai2024think} & 1) In the \textbf{think} step, the LLM reflects the navigation history and reasons about the next plans. 2) In the \textbf{act} step, the LLM predicts an action to execute a module, and the executed message is returned as context input for the next action prediction. 3) In the \textbf{ask} step, the LLM generates natural language responses to interact with the user for more information. & SR, SPL, and Success Rate weighted by the Interaction Turns (SIT). & Habitat \cite{savva2019habitat} simulator and HM3D\text{\_}v0.2 \cite{ramakrishnan2021habitat} dataset. & It does not handle broader goal types, such as image goals, or address multi-modal interactions with users in the real world.  \\ \hline
    VLFM: Vision-Language Frontier Maps for Zero-Shot Semantic Navigation \cite{yokoyama2024vlfm} & 1) It's a Zero-shot method, easily adapted or repurposed for future robotic systems performing complex tasks, provides intermediate representations that improve interpretability. & SR and SPL. & Gibson, HM3D, and MP3D datasets and Habitat simulator. & Supports single-floor episodes due to the lack of a \textbf{z} coordinate in the odometry observation. \\ 
    \hline
    RoCo: Dialectic Multi-Robot Collaboration with Large Language Models \cite{mandi2024roco} & Novel method for multi-robot collaboration that leverages LLMs for robot communication and motion planning. 2) Uses a pre-trained object detection model, \textbf{OWL-ViT}\cite{minderer2022simple}, to generate scene descriptions from top-down RGB-D camera images. & 1) Task success rate, 2) number of environment steps the agents took to succeed an episode, 3) average number of re-plan attempts at each round before an environment action is executed. & RoCoBench is built with MuJoCo\cite{todorov2012mujoco} physics engine. Text-based dataset called \textbf{RoCoBench-Text}. Real-world experiments: collaborative block sorting between a robot and a human. & 1) Assumes perception is accurate. 2) Open-loop execution: The motion trajectories from the planner are executed by robots in an open-loop fashion and lead to potential errors. \\ 
    \hline
    NavGPT: Explicit Reasoning in Vision-and-Language Navigation with Large Language Models \cite{zhou2023explicit} & 1) A fully autonomous LLM-based system tailored for navigation tasks driven by natural language instructions. 2) Equipped to interpret multi-modal inputs and unrestricted language directives, operate in open environments, and retain a record of navigational experiences.
    & TL, NE, SR, OSR, and SPL. & Mattport3D simulator \cite{anderson2018vision} and they evaluate NavGPT based on GPT-4 (OpenAI 2023)\cite{achiam2023gpt} and GPT-3.5 on the R2R dataset. & The challenges limiting LLM performance in VLN tasks mainly stem from two issues: the accuracy of verbal descriptions representing visual scenes and the effectiveness of object tracking. \\ 
    \hline
    Follow Anything (FAn): Open-Set Detection, Tracking, and Following in Real-Time \cite{maalouf2024follow} & 1) Real-time robotic system to detect, track, and follow objects in an open-vocabulary setting. Objects of interest may be specified using text, images, or clicks. 2) Leverages foundation models like CLIP \cite{radford2021learning}, DINO\cite{caron2021emerging}, and SAM\cite{kirillov2023segment} to compute segmentation masks that best align with the queried objects. & 1) mIoU and 2) the true positive detection percentage of the desired object. & Cholec80 dataset \cite{hong2020cholecseg8k} & 1) DINO+SAM yields fewer true positive detections compared to DINO-SOLO. 2) DINO+SAM provides high-quality masks once the object is detected, while DINO-SOLO masks are less refined. \\ 
    \hline
    DRAGON: A Dialogue-Based Robot for Assistive Navigation with Visual Language Grounding \cite{liu2024dragon} & 1) Uses speech to communicate with the user and a physical handle for fully autonomous navigation guidance. 2) The dialogue and navigation can be executed simultaneously. 3) If the description is ambiguous, this system will disambiguate user intents through additional dialogue. & A navigation trial is successful if the robot guides the user to the correct landmark without any delays or collisions along the route. & Dataset of 10,252 (image, question, answer) triplets to fine-tune the Visual question answering model. All the experiments are performed in the physical world. & The environment understanding modules provide limited information. \\ 
\hline
    CorNav: Autonomous Agent with Self-Corrected Planning for Zero-Shot VLN \cite{liang2024cornav} & 1) Actively adapts its plan based on feedback. If the agent receives in-plan feedback, indicating that the environmental observation aligns with the plan, it adheres to the generated plan and proceeds with the next action. 2) When faced with out-of-plan feedback, it modifies the plan accordingly. & SR, SPL, and DTS. & Develop a near-realistic simulator using \textbf{Unreal Engine 5}\cite{duan2022survey}. & 1) Relies on the outcomes of the image tagging and object detection models. 2) Models may introduce noise or miss certain objects in the environment. \\ 
\hline
    Seeing is Believing? Enhancing VLN using Visual Perturbations \cite{zhang2024seeing} & 1) Multi-Branch Architecture (MBA) extends the base model architecture by incorporating multiple branches, each of which can receive either identical or diverse visual inputs. 2) Dynamically combine the outputs of each branch to predict navigation actions based on the visual input strategies. & TL, SR, NE, SPL, RGS, and RGSPL. & REVERIE, R2R, and SOON datasets. & The SPL gains of the MBA with the optimal visual input combination method on the baselines are less significant. This may be due to the more detailed instructions in R2R imposing stricter constraints on the visual modality. \\ 
\hline
    Narrowing the Gap between Vision and Action in Navigation \cite{zhang2024narrowing} & 1) A dual-action framework for VLN-CE agents that connects high-level visual understanding with detailed spatial movements. 2) Provides the agent with the capability to choose strategic viewpoints and create corresponding low-level action plans. & SR, SPL, and nDTW. & Habitat Simulator and MP3D dataset. & While the proposed dual-action module enhances navigation at the low-level action stage, a performance disparity between high-level and low-level actions still persists.  \\ 
\hline
\end{longtable}




\section*{Statement on AI Writing Assistance}
ChatGPT was used to improve grammar and sentence clarity, with all outputs carefully reviewed and edited for relevance. ChatGPT-4o also supported the creation of realistic visualizations.

\bibliographystyle{acm}
\bibliography{references}

\end{document}